\documentclass[runningheads]{llncs}
\usepackage[T1]{fontenc}
\usepackage{booktabs}

\usepackage{url}
\usepackage{graphicx}
\usepackage{amsmath}
\usepackage{booktabs}
\usepackage[dvipsnames]{xcolor}
\usepackage{paralist}
\usepackage{dsfont}
\usepackage{bm}
\usepackage{amsfonts}
\usepackage{float}
\usepackage{ amssymb }
\usepackage{threeparttable}
\usepackage[ruled,vlined,linesnumbered]{algorithm2e}%
\usepackage[noend]{algpseudocode}
\usepackage{caption}
\usepackage{subcaption}
\usepackage{multirow}
\usepackage{pifont}

\newcommand{\sstitle}[1]{\vspace{1pt}\noindent\textbf{#1.\/}}

\newcommand{\cmark}{\ding{51}}%
\newcommand{\xmark}{\ding{55}}%

\def\Snospace~{\S{}}

\def\eqref#1{Eq.~\ref{#1}}

\def\figref#1{Fig.~\ref{#1}}
\def\tabref#1{Tab.~\ref{#1}}

\makeatletter
\newcommand{\removelatexerror}{\let\@latex@error\@gobble}
\makeatother

\begin{document}

\title{Fast-FedUL: A Training-Free Federated Unlearning with Provable Skew Resilience}

\titlerunning{Fast-FedUL}

% \author{
% First Author$^1$\footnote{Contact Author}\and
% Second Author$^2$\and
% Third Author$^{2,3}$\And
% Fourth Author$^4$\\
% \affiliations
% $^1$First Affiliation\\
% $^2$Second Affiliation\\
% $^3$Third Affiliation\\
% $^4$Fourth Affiliation\\
% \emails
% \{first, second\}@example.com,
% third@other.example.com,
% fourth@example.com

\author{Thanh Trung Huynh\inst{1}
\and
Trong Bang Nguyen\inst{2} 
\and
Phi Le Nguyen\inst{2}
\and
Thanh Tam Nguyen\inst{3}
\and
Matthias Weidlich\inst{4}
\and
Quoc Viet Hung Nguyen\inst{3}
\and
Karl Aberer\inst{1}
}
\authorrunning{T. T. Huynh et al.}
\institute{Ecole Polytechnique Federale de Lausanne, Switzerland 
\and
Hanoi University of Science and Technology, Vietnam
\and 
Griffith University, Australia
\and
Humboldt-Universitat zu Berlin
}

\maketitle

\begin{abstract}
Federated learning (FL) has recently emerged as a compelling machine learning paradigm, prioritizing the protection of privacy for training data. The increasing demand to address issues such as ``the right to be forgotten'' and combat data poisoning attacks highlights the importance of techniques, known as \textit{unlearning}, which facilitate the removal of specific training data from trained FL models.
Despite numerous unlearning methods proposed for centralized learning, they often prove inapplicable to FL due to fundamental differences in the operation of the two learning paradigms. Consequently, unlearning in FL remains in its early stages, presenting several challenges.
Many existing unlearning solutions in FL require a costly retraining process, which can be burdensome for clients. Moreover, these methods are primarily validated through experiments, lacking theoretical assurances.
In this study, we introduce Fast-FedUL, a tailored unlearning method for FL, which eliminates the need for retraining entirely. 
Through meticulous analysis of the target client's influence on the global model in each round, we develop an algorithm to systematically remove the impact of the target client from the trained model. In addition to presenting empirical findings, we offer a theoretical analysis delineating the upper bound of our unlearned model and the exact retrained model (the one obtained through retraining using untargeted clients). 
Experimental results with backdoor attack scenarios indicate that Fast-FedUL effectively removes almost all traces of the target client (achieving a mere 0.01\% success rate in backdoor attacks on the unlearned model), while retaining the knowledge of untargeted clients (obtaining a high accuracy of up to 98\% on the main task). Significantly, Fast-FedUL attains the lowest time complexity, providing a speed that is 1000 times faster than retraining.
Our source code is publicly available at \url{https://github.com/thanhtrunghuynh93/fastFedUL}.

\keywords{Machine Unlearning  \and Federated Learning \and Skew Resilience.}

\end{abstract}

\section{Introduction}
\label{sec:intro}
With the rapid advancement of AI and deep learning, there is growing awareness of potential adverse impacts, with the privacy of data used in training AI models emerging as a significant issue~\cite{nguyen2024survey}. Efforts to address this concern include various strategies, such as encrypting data~\cite{lauter2022private,lee2022privacy} or employing distributed training methods~\cite{wang2022communication,zeng2023fedlab}. Federated Learning~\cite{wang2022communication,zeng2023fedlab} emerges as a prominent solution in privacy preservation, allowing data holders to collaboratively train a model while keeping their data secure. 
Meanwhile, regulations have been established to afford data providers the authority to withdraw their supplied data~\cite{regulation2016regulation}. Unlearning~\cite{bourtoule2021machine,che2023fast,wang2022federated}, a technique in machine learning, plays a crucial role in facilitating this right by enabling the removal of specific data from a trained model. While numerous academic endeavors have focused on addressing unlearning challenges within the centralized learning framework~\cite{cha2023learning,bourtoule2021machine,nguyen2022survey}, these methods cannot be directly applied to the federated learning paradigm due to significant operational differences. As a result, unlearning in federated learning is still in its early stages. 

To fill in this gap, this study delves into the unlearning challenge within the realm of federated learning, with a specific emphasis on client-level unlearning — discarding entire data associated with one or several clients. This scenario is particularly pertinent in federated learning, where multiple clients engage in model training; thus, there are instances where specific clients may wish to retract their contributions post-participation in the federation. Moreover, some clients may exhibit malicious behavior~\cite{wang2020attack,cao2022mpaf,nguyen2023poisoning}, requiring the server to eliminate any tainted knowledge acquired from these sources.

The most straightforward approach to unlearning involves retraining from scratch with the participation of all clients except the target clients (i.e., those seeking to unlearn). Hereafter, we refer to the model obtained through this retraining process as the \textit{retrained model}.
However, this approach is not feasible due to the significant time and computational resources it demands. To this end, several solutions have been proposed~\cite{liu2021federaser,halimi2022federated,wu2022federated}, with the prevailing approach involves utilizing gradient ascent to subtract the accumulated historical updates of target clients from the trained model. 
Nevertheless, it has been noted that removing these gradients can distort the model and significantly reduce its effectiveness~\cite{wu2022federated}. Consequently, various techniques have been proposed to recalibrate the unlearned model, with the predominant method being to retrain through several iterations. In~\cite{wu2022federated}, the authors employ knowledge distillation to transfer knowledge from the old global model to the unlearned one, thereby improving its performance. FedAF~\cite{li2023federated} adopts the incremental learning paradigm to obtain new memories (excluding data from target clients), thereby overwriting old knowledge and achieving the unlearning objective. 
In~\cite{halimi2022federated,liu2022right,wang2023bfu}, the remaining clients collaborate to undergo several rounds of retraining in order to refine the unlearned model. 
It is apparent that most of the existing unlearning methods in federated learning necessitate additional training iterations, consequently adding extra burdens on clients. Moreover, a majority of them have solely undergone experimental evaluation, lacking theoretical analysis regarding the effectiveness of the unlearned model, particularly in comparison to a model retrained from scratch.

Our study introduces a federated unlearning method to address the aforementioned issues. Initially, we present a retraining-free unlearning mechanism that operates solely on clients' historical updates, eliminating the need for any retraining process. Additionally, to optimize memory usage and expedite the unlearning process, we propose an algorithm for selecting essential historical updates to store based on their significance.
Notably, in addition to presenting experimental findings demonstrating our method's efficacy, we conduct theoretical analyses to establish the upper bound of the discrepancy between our unlearned model and the exact retrained model.
The main contributions of our work are three-fold as follows.
\begin{itemize}
    \item We introduce \textit{Fast-FedUL}, a mechanism for unlearning that systematically eliminates the influence of a target client on the global model across historical training rounds. The theoretical foundation of Fast-FedUL is grounded in a comprehensive analysis of how the target client progressively impacts the global model over successive training rounds. Consequently, Fast-FedUL guarantees the complete removal of the target client's contribution and effectively alleviates distortion in the global model.
    \item We design a streamlined method for sampling and storing historical updates involves selectively retaining crucial gradients from clients during each training round. This approach allows us to conserve server memory required for storing historical updates, while simultaneously reducing the computational burden of the unlearning process.
    \item We perform theoretical analysis to set an upper bound on the disparity between the model unlearned by Fast-UL and the one obtained through retraining from scratch. Additionally, comprehensive experimental results demonstrate that Fast-UL significantly reduces execution time compared to other approaches while efficiently eliminating the knowledge influenced by the target client and retaining knowledge from the other clients.
\end{itemize}
The remainder of the paper is organized as follows.
In Section \ref{sec:related}, we introduce existing works in federated learning and unlearning, and underscore their limitations. Section \ref{sec:formulation} formulates the problem and elucidates our design principles. Details on Fast-FedUL are presented in Section \ref{sec:proposal}, followed by an evaluation of its performance in Section \ref{sec:exp}. Finally, Section \ref{sec:con} provides concluding remarks.
\section{Related Work}
\label{sec:related}
\textbf{Federated Learning.}
Federated Learning (FL) is a machine learning paradigm designed to train a model across decentralized and distributed data sources while preserving data privacy and minimizing communication overhead. The conventional FL framework comprises two main components: a server denoted as $S$ and a group of clients represented as $E=\{e_1, ..., e_N\}$. 
The FL training process involves multiple communication rounds, with a specific subset of clients participating in each round.
At the beginning of a communication round $t$, clients receive the global model $\mathcal{M}_{t-1}$ from the server. Subsequently, each client $e_i$ utilizes its private dataset $D_i$ to update the weights of $\mathcal{M}_t$. After completing local training, each client $e_i$ transmits its updates (denoted as $\Delta \mathcal{M}_t^i$) back to the server, which then aggregates and updates the global model:
\begin{equation}
    \label{eqn:fed_agg}
    \mathcal{M}_{t}=\mathcal{M}_{t-1} + Agg(\{\Delta \mathcal{M}_t^i\}),
\end{equation}
where $Agg(.)$ indicates an aggregation function.

Despite its benefits in safeguarding data privacy and harnessing distributed resources, FL remains vulnerable to adversarial attacks~\cite{xie2019dba,wang2020attack,cao2022mpaf,chang2022example}, wherein malicious clients may introduce tainted data during local training, leading to adverse impacts on the global model.
Among the diverse forms of attacks targeting FL, backdoor attacks~\cite{xie2019dba,wang2020attack} pose a significant threat due to their difficulty in detection and mitigation. In a backdoor attack scenario, malicious clients deliberately introduce corrupted data to deceive the model into producing inaccurate predictions, particularly on data exhibiting predefined characteristics chosen by the attackers (referred to as backdoor tasks), while preserving normal behavior for other data (referred to as normal tasks).

\noindent \textbf{Centralized Unlearning.}
Machine unlearning for centralized settings is the task of removing certain data and its influence from a trained model. Due to recent legal regulations such as the ``Right to be forgotten''~\cite{regulation2016regulation} and the European Union’s General Data Protection Regulation (GDPR)~\cite{voigt2017eu}, this task has gained much attention since its first introduction~\cite{cao2015towards}. Several works were to explore machine unlearning on different types of centralized settings~\cite{mehta2022deep,bourtoule2021machine,golatkar2020eternal,wu2020deltagrad,chien2022efficient}.
Methods for machine unlearning in supervised learning have been developed and can be categorized into two main strategies: implementing further fine-tuning training steps or altering the training framework to facilitate unlearning more effectively. The former can be found in~\cite{golatkar2020eternal}, where the authors utilized a reverse Newton step on a previously trained model to remove data.
 The latter approach is detailed in~\cite{bourtoule2021machine}, where Bourtoule et al. designed a generic unlearning framework named SISA. 

\noindent \textbf{Federated Unlearning}
While numerous unlearning techniques have been developed for centralized settings, they are not directly applicable to FL due to the inherent disparities between these two approaches. In contrast to the extensively studied centralized unlearning methodologies, unlearning within the context of FL has only recently garnered attention, with the pioneering work being FedEraser~\cite{liu2021federaser}.
In FedEraser, the unlearned model is rebuilt from the gradients of the clients except for the target client, while calibrating their historical gradients. A reverse learning process using  project gradient ascent was proposed in~\cite{halimi2022federated}. In general, 
both techniques attempt to reconstruct the whole model by considering all the  clients' updates and, hence, incur very high computational costs. Recently, it 
was suggested to iteratively erase the historical updates of only the target client~\cite{wu2022federated}. This removal scheme significantly speeds up the  unlearning process, but also distorts the global model. To cater for this  effect, a post-processing step was proposed that distils the knowledge of the  original model and use it to adapt the model after removal of the target client's gradients. This approach incurs overhead and 
violates the data privacy of the clients, though. The limitation of current methods lies in their requirement for retraining the unlearned model across multiple iterations, placing a substantial computational and time burden on clients.

In contrast to existing approaches, our method entirely eliminates the need for model retraining. %

\section{Problem Formulation and Design Principles}
\label{sec:formulation}
\noindent \textbf{Problem Formulation.} Let $\mathcal{M}_T$ be the global model obtained by 
federated learning over a set of clients $E$. 
Given a target client $e_u\in E$, the problem of \textit{Federated Unlearning} asks to construct a model $\mathcal{M}^{\prime}_{T}$ with minimal distance to the retrained model $\mathcal{M}^*_T$ (which is obtained by federated learning over clients $E \setminus \{e_u\}$).\\

\sstitle{Design Principles}
We argue that an ideal federated unlearning framework shall incorporate the following aspects: 
\begin{compactdesc}
    \item[C1 - Model utility:] The problem of federated unlearning is 
    bi-objective: (i) the contribution of the target client shall be removed, 
    and (ii) the utility of the model is maintained. 
    \item[C2 - Irreversible learning:] In the learning process as formulated in Eq. 1, updates of the global model depend on the previous updates of the clients, which, in turn, utilize the global model from the previous iteration. Consequently, even after removing the influence of the target client from the global model, it persists in the local models and continues to accumulate through subsequent iterations. Thus, unlike in the centralized setting, a federated unlearning technique must address the irreversible learning process.
    \item[C3 - Non-determinism:] The clients whose gradients are utilized to update the global model are randomly sampled per iteration (Eq. 1), rendering the learning process non-deterministic. Consequently, controlling unlearning and its impact on the global model becomes challenging.
    \item[C4 - Privacy:] In federated learning/unlearning, the central server is expected to not have any access to the local data of clients. Thus, any correction of the global model that is based on local data, such as knowledge distillation~\cite{wu2022federated}, violates the clients' data privacy.
    \item[C5 - Efficiency:] The unlearning process needs to be efficient, so that retraining the global model without the target client's data is not feasible. Also, as the computing power of the clients is usually limited, the unlearning process should be conducted on the server rather than on the clients as \cite{liu2021federaser}.       \\ 
\end{compactdesc}

\noindent Existing techniques inadequately address these requirements. FedEraser \cite{liu2021federaser} retrains the model from retained updates with a calibration method to expedite the process. However, this calibration alters the order (violating C3) and only slightly improves the time compared to retraining from scratch (violating C5). CDP-FedUL \cite{wang2022federated} achieves unlearning by pruning entire information classes from the global model, which is strict and incapable of handling sophisticated backdoor attacks that inject backdoor data into existing classes (violating C1). PGA-FedUL \cite{halimi2022federated} attempts to reverse the learning process using projected gradient ascent, disregarding the irreversible nature of learning (violating C2) and consequently undermining model quality (violating C1). KD-FedUL \cite{wu2022federated} eliminates historical gradients and performs knowledge distillation to rectify model skew. However, this step necessitates additional training (violating C5) and direct access to clients' data (violating C4).
We compare the novelty of our approach against the SOTAs in \tabref{tab:functionality}.
\begin{table}[bt]
    \centering
      \caption{Functionality comparison to existing techniques.} 
    \label{tab:functionality}
   \resizebox{0.6\linewidth}{!}{%
      \begin{tabular}{l c c c c c}
    \hline
    & (\textbf{C1})  & (\textbf{C2}) & (\textbf{C3}) & (\textbf{C4}) & (\textbf{C5}) \\ 
    \hline
    Retrain & \cmark  & \cmark & \cmark & \cmark & \xmark \\ 
    FedEraser \cite{liu2021federaser} \quad & \cmark  & \cmark & \cmark & \cmark & \xmark \\     
    CDP-FedUL \cite{wang2022federated} \quad & \xmark  & \cmark & \xmark & \cmark & \cmark \\
    PGA-FedUL \cite{halimi2022federated} \quad & \xmark  & \xmark & \xmark & \cmark & \cmark \\ 
    KD-FedUL \cite{wu2022federated} \quad & \cmark  & \cmark & \cmark & \xmark & \xmark \\ 
    \hline
    \textbf{Fast-FedUL (Ours)} & \cmark  & \cmark & \cmark & \cmark & \cmark \\ 
 \hline
    \end{tabular}%
 } 
\end{table}

\section{Fast-FedUL}
\label{sec:proposal}
This section presents the details of Fast-FedUL, our proposed federated unlearning method. 
We start with an overview of Fast-FedUL in Section \ref{sec:overview}. Subsequently, we present our algorithm for sampling crucial updates in Section \ref{sec:storage}. Finally, we delve into the details of unlearn algorithm in Sections \ref{sec:removal} and provide theoretical analysis of Fast-FedUL in Section \ref{sec:theoretical-analysis}. 
\vspace{-5pt}
\subsection{Overview}
\label{sec:overview}
\begin{figure*}[tb]
	\centering
    \includegraphics[width=0.9\linewidth]{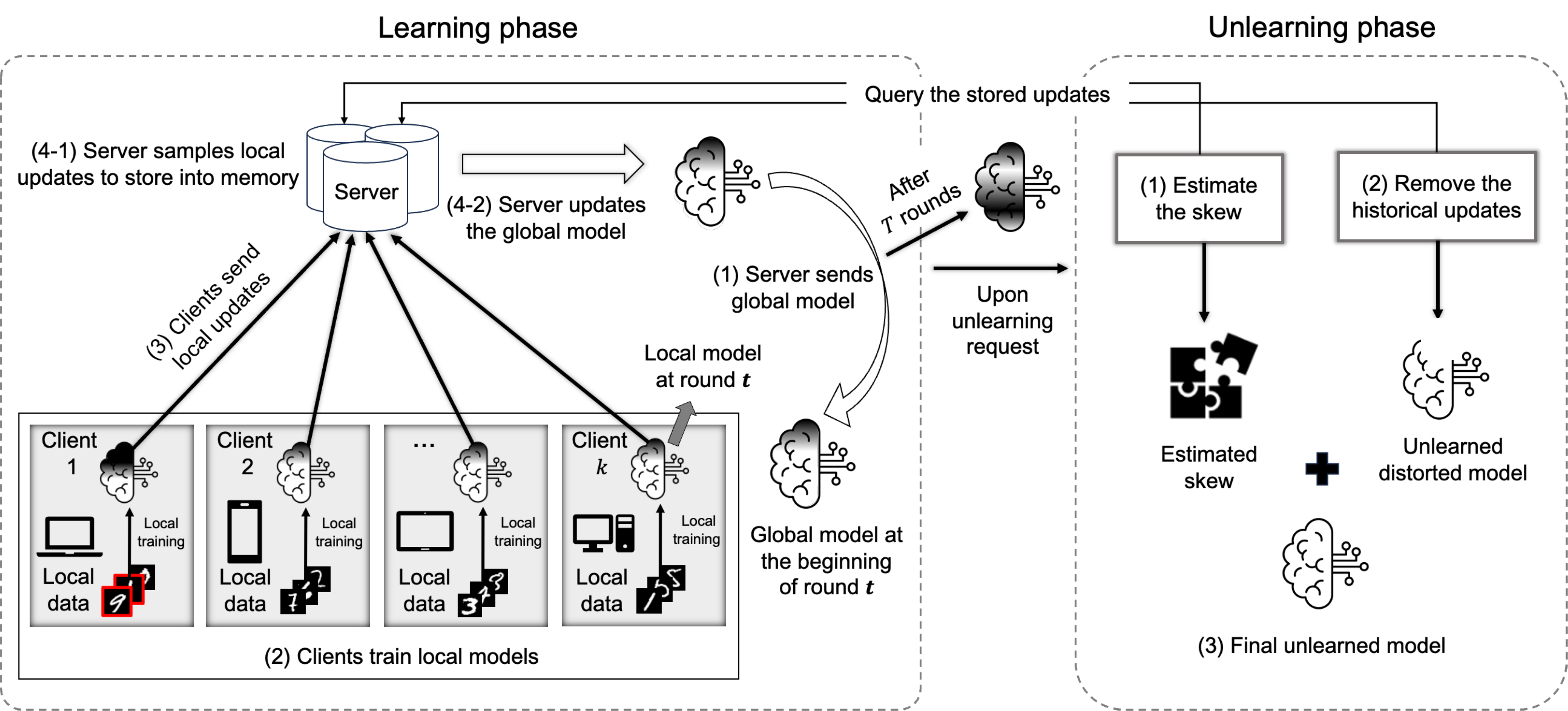}
    \caption{\textbf{Overview of Fast-FedUL.} During the learning phase, the server selectively stores the sampled local updates. Upon receiving an unlearning request, the server retrieves the stored updates from memory and performs reverse subtraction using the estimated skew.}
    \label{fig:framework}
    \vspace{-10pt}
\end{figure*}
Considering the design principles mentioned in the previous section, we introduce Fast-FedUL, a novel federated unlearning framework as shown in Fig. \ref{fig:framework}. 
Fast-FedUL is agnostic to the underlying Federated Learning model and can be seamlessly applied  any FL architectures. Its unlearning mechanism encompasses two primary stages: (1) \textit{Skew estimation} and (2) \textit{Removal of updates from the target client}. Upon receiving the unlearning request, the server initiates our proposed skew estimation algorithm to gauge the target client's impact on the global model in every round. Subsequently, armed with this estimation, the server retraces the historical updates and systematically eliminates the target client's influence from the global model. 
In addition, to minimize storage requirements for historical updates and reduce computational costs during the unlearning process, we propose an algorithm that assists the server in selecting and retaining only significant updates during the training process.

Unlike conventional methods that merely subtract the target client's historical gradients, Fast-FedUL capitalizes on the estimated skew to precisely eradicate the target client's contribution, thereby preserving its utility (C1).Due to the irreversible nature of federated learning (\textbf{C2}), exact model skew calculation is infeasible. Hence, we estimate it using Lipschitz conditions per client and iteration, aggregating to approximate the overall skew. Our approach applies per-client, per-iteration, enabling precise unlearning only for sampled clients (\textbf{C3}). The estimated skew is directly removed from the global model, preserving its utility, obviating post-processing (\textbf{C4} and \textbf{C5}). For brevity, detailed procedures for federated learning and unlearning are described in the Supplementary. 

\vspace{-5pt}
\subsection{Efficient Sampling of Local Updates}
\label{sec:storage}
To facilitate the later unlearning process, which involves undoing past updates, the historical updates are required to be stored during the learning process. We propose a sampling algorithm that selectively aggregates and stores only the significant updates, thereby optimizing storage costs and expediting the unlearning process.

\noindent Intuitively, at each training round $t$, our method chooses a subset $C^t \subseteq E$ comprising of $m$ clients (where $m$ is a predetermined parameter) in such a way that the aggregated updates of these $m$ clients closely resemble the aggregated updates of all clients.
Without loss of generality, we assume that the system uses the popular mean aggregator \textit{FedAvg}~\cite{mcmahan2017communication}, then the objective can be mathematically represented by
\begin{equation}
    \label{eqn:obj1}
    \operatorname{Minimize} ||  Agg'_{e_j \in C^t}(\Delta \mathcal{M}_t^j )  - \sum_{e_i \in E}{\Delta \mathcal{M}_t^i}||^2,
\end{equation}
where $Agg'_{e_j \in C^t}(\Delta \mathcal{M}_t^i)$ denotes the aggregation of the sampled updates. 
To mitigate bias towards frequently selected updates, we employ weighted sum to aggregate updates from the sampled clients. Accordingly, the gradient aggregation over the sampling $C^t$ can now be written as: 
\begin{equation}
    \label{eqn:obj2}
    Agg'_{e_j \in C^t}(\Delta \mathcal{M}_t^j) := \sum_{e_j \in C^t}{ \frac{1}{p_t^j}\Delta \mathcal{M}_t^j},
\end{equation}
where $p_t^j$ represents the probability that a client $e_j$ is sampled in round $t$.
Consequently, the objective function now can be expressed by:
\begin{equation}
    \label{eqn:obj3}
    \operatorname{Minimize} || \sum_{e_j \in C^t}{ \frac{1}{p_t^j}\Delta \mathcal{M}_t^j} - \sum_{e_i \in E}{\Delta \mathcal{M}_t^j}||^2. 
\end{equation}
To solve such an optimization problem,  we adopt the partial participation framework proposed in \cite{horvath2019nonconvex}.
This framework associates a sampling strategy $C^t$ with a probability matrix $\mathbf{P}_t \in \mathbb{R}^{N \times N}$, where each entry $\mathbf{P}_t(i,j)$ is defined by the probability for both two clients $e_i$ and $e_j$ being selected by $C^t$, and a diagonal entry $\mathbf{P}_t(i,i)$ represents the probability for $e_i$ being sampled ($\mathbf{P}_t(i,i) = p_t^i$).

\noindent By applying a lemma from \cite{horvath2019nonconvex} (see Lemma 1 in the Supplementary) we derive:
\begin{equation}
\label{eqn:obj4}
  \mathbb{E}  ( || \sum_{e_j \in C^t}{ \frac{1}{p_t^j}\Delta \mathcal{M}_t^i} - \sum_{e_i \in E}{\Delta \mathcal{M}_t^i} ||^2  ) \leq \sum_{e_i \in E} \frac{v_i}{p_t^i}\left\|\Delta \mathcal{M}_t^i\right\|^2,
\end{equation}
where $v = [v_1, \dots, v_N]$ is a vector satisfying Condition (1) in Lemma 1 (Appendix). 
It turns out that any satisfied vector $v$ must hold that $\forall i, v_i \geq 1 - p_t^i$ (see Proof 1 in the Supplementary). 
By applying this condition to \eqref{eqn:obj4}, we gain a tighter upper bound: 
\small
\begin{equation}
\label{eqn:obj5}
  \mathbb{E} ( || \sum_{e_j \in C^t}{ \frac{1}{p_t^j}\Delta \mathcal{M}_t^i} - \sum_{e_i \in E}{\Delta \mathcal{M}_t^i} || ^2 ) \leq \sum_{e_i \in E} \frac{1 - p_t^i}{p_t^i}\left\|\Delta \mathcal{M}_t^i\right\|^2.
\end{equation}
\normalsize
The upper bound on the right side can be seen as an effective estimation of the difference on the left side.
Thus, our problem can now be approximated by minimizing the right-hand side of Eq. \ref{eqn:obj5}.
Intuitively, the minimum can be attained when $p_t^i$ is proportional to $||\Delta \mathcal{M}_t^i||^2$. 
When combined with the cardinality condition of $C^t$ being equal to $m$, we deduce the optimal solution as follows:
\small
\begin{equation}
\label{eqn:probability}
p_t^i=\left\{\begin{array}{ll}
\frac{\left(m+l-N)*\|\Delta \mathcal{M}_t^i\right\|}{\left.\sum_{j=1}^l \|\Delta \mathcal{M}_t^{(j)}\|\right.}, & \text {if} \left\|\Delta \mathcal{M}_t^i \right\| <\left\|\Delta \mathcal{M}_t^{(l+1)}\right\|\\
1, & \text {otherwise} 
\end{array},\right.    
\end{equation}
\normalsize
where $\|\Delta \mathcal{M}_t^{(j)}\|$ denotes the $j$-th smallest value in $\left\{\left\|\Delta \mathcal{M}_t^i\right\|\right\}_{e_i \in E}$ and $l$ is the largest integer satisfying $0<m+l-N \leq \sum_{i=1}^l\left\|\Delta \mathcal{M}_t^{(i)}\right\| /\left\|\Delta \mathcal{M}_t^{(l)}\right\|$.
Detailed proofs are provided in Appendix A.4 (Supplementary material).
 
\subsection{Removal of Target Client Updates}
\label{sec:removal}
Let $\mathcal{M}_T$ denote the final global model, and $e_u$ represent the target client seeking unlearning. Our objective is to trace back and deduct all influence of $e_u$'s historical updates $\Delta \mathcal{M}_u^t$ ($\forall t \in [1, T]$) from $\mathcal{M}_T$.

\noindent Let $\mathcal{M}^*_T$ depict the model obtained by retraining from scratch with all clients excepting $e_u$. Then we seek to estimate the difference $\Delta_T$ between $\mathcal{M}^*_T$ and $\mathcal{M}_T$:
\begin{equation}
\label{eqn:final_model}
    \Delta_T = \mathcal{M}^*_T - \mathcal{M}_T.
\end{equation}
Denote $\mathcal{M}_{t}^{*}$ as the retrained model at a round $t$, then $\mathcal{M}_{t}^{*}$ and $\mathcal{M}_{t}$ can be recursively represented as follows:
\small
\begin{align}
\label{eqn:clean_update}
\mathcal{M}_{t}^{*} &= \mathcal{M}_{t-1}^{*} + \frac{1}{N-1} \sum_{e_i \in C_F^{t-1}} \Delta 
{\mathcal{M}^{*}}^i_{t-1},\\
\label{eqn:dirty_update}
\mathcal{M}_{t} &= \mathcal{M}_{t-1} + \frac{1}{N} \sum_{e_i \in C^{t-1}} \Delta \mathcal{M}^i_{t-1},
\end{align}
\normalsize
where $C_F^t = C^t \setminus \{e_u \}$ is set of sampled clients excluding the target client. $\Delta_t = \mathcal{M}_{t}^{*}-\mathcal{M}_{t}$ is hence obtained by subtracting \eqref{eqn:dirty_update} from \eqref{eqn:clean_update}:
\small
\begin{equation}
\label{eqn:actual_delta}
\Delta_{t} =  \Delta_{t-1} + \frac{1}{(N-1)} \sum_{e_i \in C_F^{t-1}} 
\epsilon^i_{t-1} 
 +
\frac{1}{N(N-1)}\sum_{e_i \in C_F^{t-1}} \Delta \mathcal{M}^i_{t-1} - \frac{1}{N} \Delta \mathcal{M}^u_{t-1},
\end{equation}
where $\epsilon^i_t = \Delta \mathcal{M}_{t}^{*i} - \Delta \mathcal{M}_{t}^{i}$ represents the local gradient skew induced by client $e_i$ in round $t$.
The most critical challenge now is to estimate $\epsilon^i_t$. Computing this skew from scratch is resource-intensive since it entails involving all clients in each iteration. To address this, we propose an efficient estimation method for this term which will be presented in Section \ref{sec:skew_estimation}.
Now, by substituting our estimated skew $\epsilon^i_t$ (Eq. \ref{eqn:approx_epsilon}) into Eq. \ref{eqn:actual_delta} we obtain the following recursive formulation:
\small
\begin{align}
\label{eqn:recur_equa}
\Delta_t \approx (1 + \alpha)\Delta_{t-1} 
  + \frac{1}{N(N-1)} \sum_{e_i \in 
C_F^{t-1}} \Delta \mathcal{M}^i_{t-1} - \frac{1}{N} \Delta \mathcal{M}^u_{t-1}.
\end{align}
\normalsize
Using this formula, the unlearning process can efficiently construct the final model difference $\Delta_T$, which is then used to derive the unlearned model $\mathcal{M}'_T$ by \eqref{eqn:final_model}. The whole unlearning process is summarized in Appendix A.2. 
\vspace{-5pt}
\subsection{Skew Estimation}
\label{sec:skew_estimation}
In this section, we present our algorithm to estimate the local gradient skew $\epsilon_t^i$ caused by each client $e_i$ in round $t$.
We have:
\small
\begin{multline}
\label{eqn:gradients}
|\epsilon^i_t| = | \Delta {\mathcal{M}^{*}}^i_t - \Delta \mathcal{M}^i_t |
= 
|\frac{\partial \mathcal{L}(f(X_{i}, \mathcal{M}_{t}^{*}), Y_{i})}{\partial \mathcal{M}_{t}^{*}} - 
\frac{\partial \mathcal{L} ( f( X_{i}, \mathcal{M}_{t} ), Y_{i} )}{\partial \mathcal{M}_{t}} |,
\end{multline}
\normalsize
where $\mathcal{L}$ is the loss function of the original task, $(X_i, Y_i) \in D_i$ are the samples and labels from the local data set $D_i$, and $\frac{\partial \mathcal{L}(f(X_{i}, \theta), Y_{i})}{\partial \theta}$ is the gradient of the loss function $\mathcal{L}$ over the variable $\theta$. As $(X_i, Y_i)$ is constantly specified for the client $e_i$, the gradient can be seen as a function of $\theta$, denoted as $F_i(\theta)$.  $\Delta {\mathcal{M}^{*}}^i_t$ and $\Delta \mathcal{M}^i_t$ are values of the same function $F_i$ over the variables $\mathcal{M}_{t}^{*}$ and $\mathcal{M}_{t}$, respectively. Assume that the function $F_i$ is uniformly continuous and strong convex, we have the inequation of Lipschitz continuous condition for $F_i$:
\begin{align}
\label{eqn:Lipschitz}
\nonumber
|\epsilon^i_t| = |F_i(\mathcal{M}_{t}^{*}) - F_i(\mathcal{M}_{t})| \leq \mathcal{K} | \mathcal{M}_{t}^{*}-\mathcal{M}_{t} |,
\end{align}
where $\mathcal{K}$ is the Lipschitz constant. Note that this inequation holds for every 
local skew $\epsilon^i_t$. Thus, we can establish the bound for accumulated 
local skew $\epsilon_t$ of iteration $t$ using the difference between 
$\mathcal{M}_{t}^{*}$ and $\mathcal{M}_{t}$:
\begin{equation}
\epsilon_t = \frac{1}{N-1} \sum_{e_i \in C_F^t} \epsilon^i_t \leq \mathcal{K} | \mathcal{M}_{t}^{*}-\mathcal{M}_{t} |.
\end{equation}
Based on the bound, we can use a hyperparameter $\alpha 
\in [-\mathcal{K},\mathcal{K}]$ to approximate $\epsilon^i_{t}$: $\epsilon^i_{t} \approx \alpha*\Delta_{t}$; referred to as Lipschitz coefficient hyperparameter. Accumulating over the clients, we obtain an estimation for the model skew $\epsilon_t$ of each round $t$: 
\begin{equation}
\label{eqn:approx_epsilon}
\epsilon_t = \frac{1}{N-1} \sum_{e_i \in C_F^t} \epsilon^i_t \approx \alpha \Delta_t.
\end{equation}

\vspace{-25pt}
\subsection{Theoretical Analysis}
\label{sec:theoretical-analysis}
\textbf{Theorem 1.} The difference between $\mathcal{M}^{\prime}_T$, the model unlearned by Fast-FedUL, and $\mathcal{M}^{*}_T$, the model retrained from scratch, is bounded as follows:
\begin{equation}
\left \| \mathcal{M}^{\prime}_T - \mathcal{M}^{*}_T \right | \leq   \left ( \mathcal{K} + |\alpha| \right ) \sum_{j=0}^{T-2} \frac{(1 + \mathcal{K})^{T-1-j} - |1+\alpha|^{T-1-j}}{(1 + \mathcal{K}) - |1+\alpha|} \|\gamma_{j}\|,
\end{equation}
where $\gamma_{i}=\frac{1}{N(N-1)}\displaystyle\sum_{e_i \in C_F^{t-1}}\Delta \mathcal{M}^i_{t-1} - \textstyle\frac{1}{N} \Delta \mathcal{M}^u_{t-1}$.
\\
\textbf{Proof.} Let us denote by $\Delta'_{t}$ and $\Delta_t$ the disparities from the unlearned model $\mathcal{M}'_{t}$ and the retrained model $\mathcal{M}^{*}_{t}$ to the global model $\mathcal{M}_{t}$ at round $t$.
From \eqref{eqn:actual_delta} and recursive formula for $\Delta'_{t}$ similar to \eqref{eqn:recur_equa} (Algo. 1 in Appendix), we have:
\begin{eqnarray}
\nonumber    \|\mathcal{M}'_{t} - \mathcal{M}^{*}_{t}\| = \|\Delta'_{t}-\Delta_{t}\| & = & \|(1+\alpha)(\mathcal{M}'_{t-1}-\mathcal{M}^{*}_{t-1}) + \alpha \Delta_{t-1} - \epsilon_{t-1}\| \\
\nonumber     &\leq &|1+\alpha|\|\mathcal{M}'_{t-1}-\mathcal{M}^{*}_{t-1}\| + |\alpha|\|\Delta_{t-1}\| + \|\epsilon_{t-1}\| \\
  \label{eq:proof1}  &\leq &|1+\alpha|\|\mathcal{M}'_{t-1}-\mathcal{M}^{*}_{t-1}\| + (\mathcal{K} + |\alpha|)\|\Delta_{t-1}\|.
\end{eqnarray}
By aggregating Eq. \ref{eq:proof1} over all values of $t$ ranging from $0$ to $T$, we derive: 
\begin{align}
    \label{eqn:diff_unlearn}
     \|\mathcal{M}'_{T} - \mathcal{M}^{*}_{T}\| \leq (\mathcal{K} + |\alpha|)\sum_{i=1}^{T-1} |1+\alpha|^{T-1-i}\|\Delta_{i}\|.
\end{align}
Meanwhile, from \eqref{eqn:actual_delta} we also have:
\begin{multline}
    \label{eqn:bound_delta}
    \|\Delta_{t}\| = \|\Delta_{t-1} + \epsilon_{t-1} + \gamma_{t-1}\|
    \leq \|\Delta_{t-1}\| + \|\epsilon_{t-1}\| + \|\gamma_{t-1}\| \\
    \leq (1 + \mathcal{K})\|\Delta_{t-1}\| + \|\gamma_{t-1}\| 
    \leq \sum_{i=0}^{t-1} (1 + \mathcal{K})^{t-1-i} \|\gamma_{i}\|.
\end{multline}
By combining \eqref{eqn:diff_unlearn} and \eqref{eqn:bound_delta}, we derive:
\begin{eqnarray}
\nonumber     \|\mathcal{M}'_{T} - \mathcal{M}^{*}_{T}\| &\leq& (\mathcal{K} + |\alpha|)\sum_{i=1}^{T-1} |1+\alpha|^{T-1-i}\sum_{j=0}^{i-1} (1 + \mathcal{K})^{i-1-j} \|\gamma_{j}\| \\
\nonumber     & = & (\mathcal{K} + |\alpha|)\sum_{j=0}^{T-2} \sum_{i=j+1}^{T-1} |1+\alpha|^{T-1-i} (1 + \mathcal{K})^{i-1-j} \|\gamma_{j}\| \\
\nonumber     & = & (\mathcal{K} + |\alpha|)\sum_{j=0}^{T-2} \frac{(1 + \mathcal{K})^{T-1-j} - |1+\alpha|^{T-1-j}}{(1 + \mathcal{K}) - |1+\alpha|} \|\gamma_{j}\| \quad \blacksquare 
\end{eqnarray}

\sstitle{Theorem 2} The time complexity of Fast-FedUL is $O(T \times N)$.

\noindent \textbf{Proof.} One of the key advantages of Fast-FedUL is that our technique is only based on recursive equation \eqref{eqn:recur_equa} and does not do any training steps. It is observable that during each instance of recursion, there are $N$ operations of addition and a single multiplication operation. This process is repeated $T$ times to compute $\Delta_T$. Consequently, the aggregate complexity of this algorithm is $O(T \times (N+1))$, which is approximately equivalent to $O(T \times N)$. 

\noindent Compared to the existing works, unlearning strategies often relies on speeding up retraining \cite{liu2021federaser} or performing post-training to recover the model utility \cite{halimi2022federated}. Such techniques need time of $O(E\times B \times T_{train})$ for each local unlearning step at client and $O(T \times N \times E \times B \times T_{train})$-complexity for the whole unlearning process, where $T,E,N,B, T_{train}$ are the number of iterations, epochs, clients, batches and time for forward and backward process, respectively. Some other algorithms only perform additional training locally without performing the FL training rounds \cite{wu2022federated}, these algorithms requires complexity of $O(|C^t_F| \times E\times B \times T_{train}) \approx O(N \times E \times B \times T_{train})$. In practical situations, for ensuring the efficiency of algorithms based on training, their hyper-parameters almost satisfy $E \times B \times T_{train}>>T$ and it can be easily seen that our technique requires significantly less computation complexity than the state-of-the-arts, which is confirmed in the experiments. 
\section{Empirical Evaluation}
\label{sec:exp}
\sstitle{Datasets}  We conduct experiments on three datasets, which vary in terms of data modality. The first two datasets are 
MNIST \cite{mnist} and CIFAR10 \cite{cifar}, which denote important benchmark 
datasets for every image classifier. The datasets consist of 60,000 images 
of 0-9 digits and 60,000 images of objects in 10 classes, respectively. We 
further use the OCTMNIST dataset from the MedMNISTv2 
repository~\cite{medmnistv2}, a MNIST-like collection of 
109,309 biomedical images. 

\sstitle{Baselines} We compare our proposed method \textit{Fast-FedUL} with five other baselines:
\begin{compactenum}
\item \emph{Retrained:} implements the naive solution that retrains the global 
model from scratch without the target client. 
\item \emph{FedEraser:} reconstructs the 
model after unlearning by retraining with a method to calibrate the retained 
updates of the clients~\cite{liu2021federaser}.
\item \emph{CDP-FedUL:} borrows the TF-IDF concept from NLP to quantize the class discrimination of channels, then prune the ones with higher degree to unlearn the target information category \cite{wang2022federated}.
\item \emph{PGA-FedUL:} is a federated unlearning technique that attempts to reverse the learning process by using projected gradient ascent \cite{halimi2022federated}.
\item \emph{KD-FedUL:} unlearns the target client from the global model by 
first removing its historical gradients, then distilling the knowledge from 
the original model to fix the skew in the model~\cite{wu2022federated}. 
\end{compactenum} 
\sstitle{Evaluation using backdoor attacks}
We assess the efficacy of the unlearning techniques through backdoor attack scenarios. The choice of the backdoor attack as the test scenario stems from its ability to provide a clear quantitative evaluation of unlearning methods based on two criteria: removing the knowledge of the target client and retaining the knowledge of untargeted clients.
In a backdoor attack, the target client injects poisoned data containing specific backdoor patterns to manipulate the global model. The objective is to induce the global model to produce incorrect behavior for input data with the backdoor patterns (backdoor task $\mathcal{T}_b$), while maintaining normal behavior for other inputs (normal task $\mathcal{T}_m$).
To counter a backdoor attack, an unlearning process must cleanse the infected global model of the contributions from the target client (i.e., the malicious client). Essentially, unlearning aims to reduce the success rate of the attack (accuracy on $\mathcal{T}_b$), while restoring the performance for the main task.
Specifically, the lower the accuracy of the backdoor task, the more effectively the target client's influence is removed from the unlearned model. Conversely, the higher the accuracy of the main task, the more knowledge of untargeted clients remains in the unlearned model.
We employ two different backdoor attack scenarios: edge-case backdoor \cite{wang2020attack} and pixel backdoor \cite{xie2019dba}. 

\sstitle{Metrics} Using the above setting of a backdoor attack, we evaluate the 
quality of the final global model after 
unlearning based on its accuracy on the \textit{main 
task} and the \textit{backdoor task}, respectively. The results for both tasks 
are compared for of each 
model after unlearning and the original model. To assess the computational 
efficiency, we measure the total runtime for unlearning in 
seconds.

\vspace{-8pt}
\subsection{End-to-end Comparison}
\label{sec:exp_accuracy}
We report an end-to-end comparison of our method (Fast-FedUL) and the baselines on the MNIST, CIFAR10, and OCTMNIST datasets. 

\begin{figure}[tb]
 \centering
 \begin{minipage}{0.42\linewidth}
\centering
    \includegraphics[width=1\linewidth]{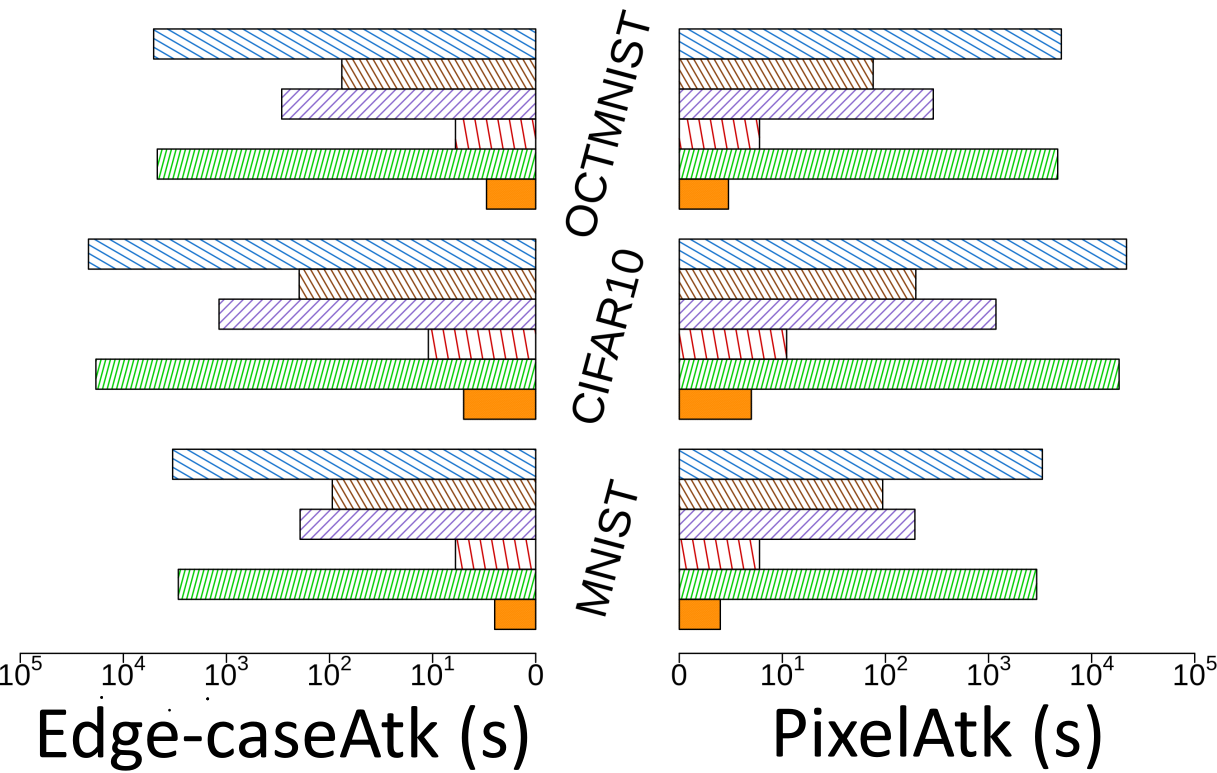}
    \caption{Unlearning time.}
    \label{fig:unlearn_time}
  \end{minipage}
  \quad
  \begin{minipage}{.408\linewidth}
\centering
    \includegraphics[width=1\linewidth]{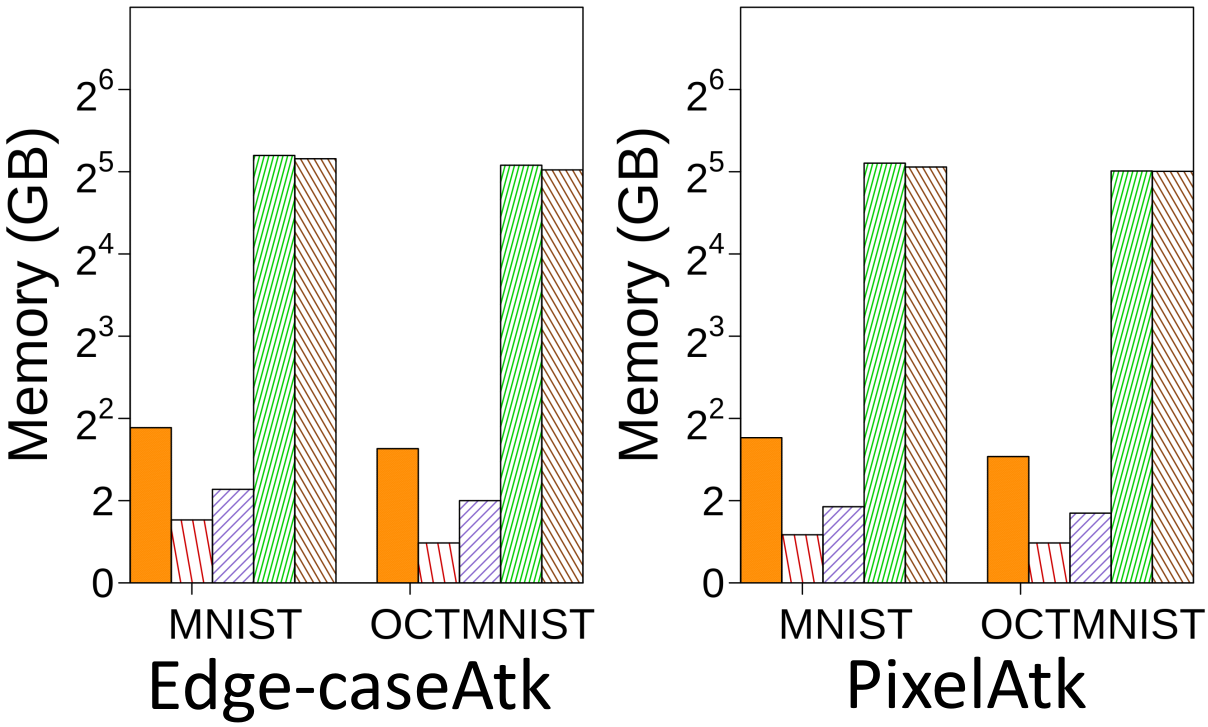}
    \caption{Memory usage.}
    \label{fig:unlearn_memory}
  \end{minipage}
  \quad
  \begin{minipage}{.1\linewidth}
\centering
    \includegraphics[width=1\linewidth]{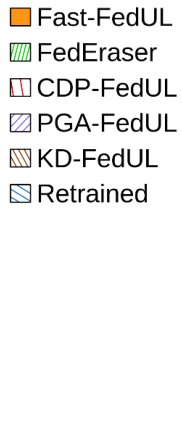}
  \end{minipage}
   \vspace{-10pt}
\end{figure}

\begin{figure}[tb]
    \centering
    \begin{subfigure}{0.35\linewidth}
    \centering
    \includegraphics[width=\linewidth]{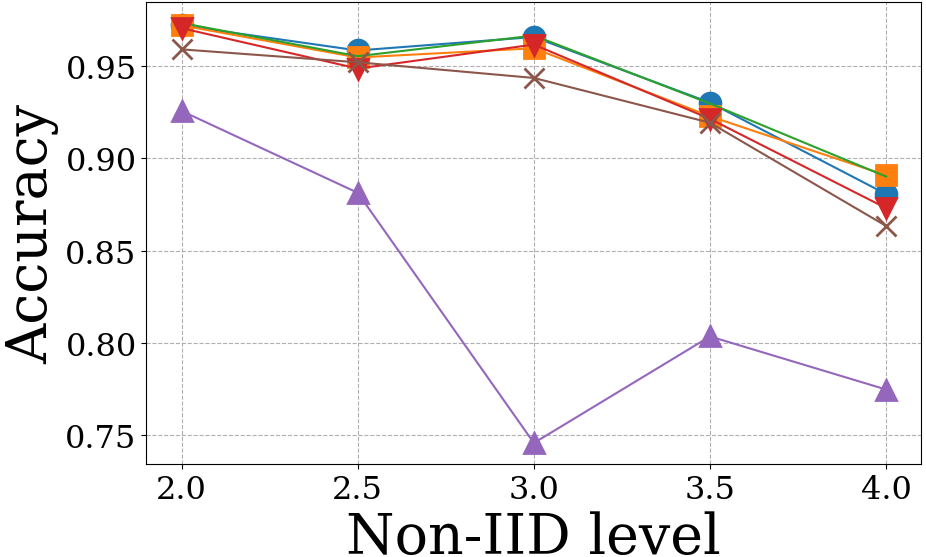}
    \caption{Main task $\mathcal{T}_m$}
    \end{subfigure}
    \hfill%
    \begin{subfigure}{0.35\linewidth}
    \centering
    \includegraphics[width=\linewidth]{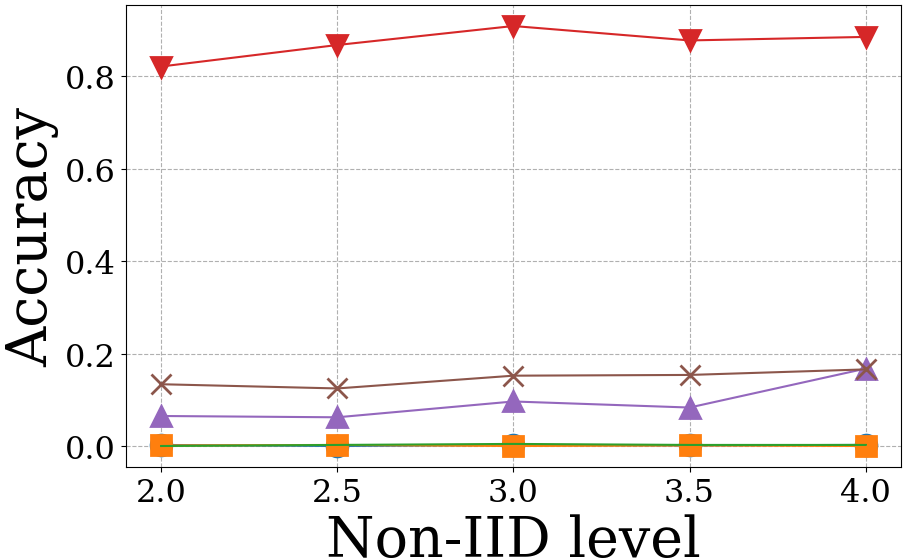}
    \caption{Backdoor task $\mathcal{T}_b$}
    \end{subfigure}
    \hfill%
    \begin{subfigure}{0.2\linewidth}
    \centering
    \includegraphics[width=\linewidth]{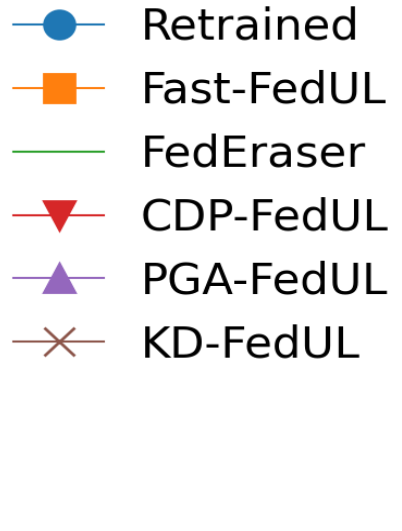}
    \end{subfigure}
    \caption{Robustness against non-IID (\textit{edge-case backdoor}).}
    \label{fig:edge_nonIID}
    \vspace{-8pt}
\end{figure}

\sstitle{Efficiency} We assess the efficiency of the techniques by presenting the execution time (Fig. \ref{fig:unlearn_time}) and memory usage (Fig. \ref{fig:unlearn_memory}) for the two attack scenarios, i.e., edge-case and pixel. In terms of execution time, our technique outperforms all other unlearning methods. Specifically, the execution time of Fast-FedUL is only a fraction of the other methods, i.e., 1/2, 1/26, 1/110, and 1/1600 of CDP-FedUL, KD-FedUL, PGA-FedUL, and FedEraser, respectively. Moverover,  Fast-FedUL is 1000 times faster than retraining model from scratch.  
Among the baselines, CDP-FedUL is the fastest, although it compromises unlearning quality for the sake of speed (see \tabref{tab:end-to-end} and \figref{fig:edge_nonIID}). 

Regarding memory usage, our technique requires significantly less memory compared to KD-FedUL and FedEraser, despite all three methods involving storing historical updates. This efficiency is attributed to our streamlined sampling strategy. 
Although PGA-FedUL and KD-FedUL save more memory than Fast-FedUL, it is worth noting that they lag far behind in both running time and accuracy. 
\begin{table}[tb]
\centering
\caption{End-to-end comparison concerning the \textbf{edge-case backdoor}. \label{tab:end-to-end}}
\scriptsize
\setlength\tabcolsep{3pt}
\resizebox{\linewidth}{!}{
\begin{threeparttable}
\begin{tabular}{@{}l|cc|cc|cc@{}}
\toprule
\multicolumn{1}{c|}{\multirow{2}{*}{Method}} & \multicolumn{2}{c|}{MNIST} & \multicolumn{2}{c|}{CIFAR-10} & \multicolumn{2}{c}{OCTMNIST} \\ \cmidrule(lr){2-7} 
\multicolumn{1}{c|}{} & \multicolumn{1}{c}{Main($\uparrow$)} & \multicolumn{1}{c|}{Backdoor($\downarrow$)} & \multicolumn{1}{c}{Main($\uparrow$)} & \multicolumn{1}{c|}{Backdoor($\downarrow$)} & \multicolumn{1}{c}{Main($\uparrow$)} & \multicolumn{1}{c}{Backdoor($\downarrow$)} \\  \midrule
\textit{Pre-unlearned} & 0.9833 & 0.7340 & 0.9773 & 0.8724 & 0.8966 & 0.9711 \\
\textit{Retrained} & 0.9846 & 0.0000 & 0.9640 & 0.0009 & 0.8990 & 0.0040 \\ \midrule
FedEraser & 0.9757 & 0.0027 & 0.9591 & 0.0050 & 0.8951 & 0.0040 \\
CDP-FedUL & 0.9808 & 0.7287 & 0.9577 & 0.7612 & 0.8959 & 0.9664 \\
PGA-FedUL & 0.8320 & 0.0080 & 0.8089 & 0.0638 & 0.8435 & 0.0000 \\
KD-FedUL & 0.9792 & 0.0319 & 0.9583 & 0.0261 & 0.8721 & 0.0141\\ \midrule
\textbf{Fast-FedUL} & 0.9737 & 0.0027 & 0.9546 & 0.0061 & 0.8817 & 0.0047 \\
\bottomrule
\end{tabular}
\end{threeparttable}
}
\vspace{-15pt}
\end{table}

\sstitle{Accuracy}
We evaluate the unlearning methods based on accuracy in both the main task and the backdoor task across two client data settings: IID and non-IID.
Due to space constraints, we only provide results for the edge-case backdoor attack. Similar results are obtained for \textit{pixel backdoor attacks} and are described in Appendix A.3 (Supplementary material). 

The results pertaining to the IID setting are presented in (\tabref{tab:end-to-end}). 
As demonstrated, Fast-FedUL effectively unlearns the target client and mitigates the backdoor attack while maintaining the model's accuracy on the main task. 
After unlearning with Fast-FedUL, the global model still achieves an accuracy of 97.37\%, 95.46\%, and 88.17 \% concerning the main task on the three datasets. 
On average, this is equivalent to 98.3\% of the model accuracy before unlearning, and nearly identical to the quality of the model retrained from scratch. Also, our technique effectively removes the threat from the attack, with the success attack rate being less than 0.01\% for all the three datasets. 
Among other methods, FedEraser and PGA-FedUL also demonstrate the capability to unlearn and counteract the backdoor attack. However, it is noteworthy that while their backdoor accuracy is comparable to that of Fast-FedUL, they demand over 100 times the computational time compared to Fast-FedUL, due to the necessity of a retraining process. 
CDP-FedUL, which excels in terms of memory usage, exhibits the poorest performance in removing the target client's knowledge, resulting in the highest accuracy in the backdoor task.

To further investigate the performance of the unlearning methods concerning the accuracy, we perform experiments with non-IID data sampled from the MNIST dataset (using the Dirichlet distribution \cite{li2022federated}). The sampling is modulated to vary the ratio between the classes that appeared most often and the least often (named as \textit{non-IID level}) from 2 to 4. 
The results in \figref{fig:edge_nonIID} demonstrate that all techniques experience reduced main task model quality as non-IID levels increase. Notably, Fast-FedUL and FedEraser stand out as the best methods, fully eliminating the backdoor attack while maintaining retraining-level model quality. In contrast, PGA-FedUL is highly vulnerable to non-IID scenarios, exhibiting a significant main task accuracy drop at a ratio of 4 due to its susceptibility to data class imbalance stemming from the projected gradient ascent process. As expected, CDP-FedUL performs poorly in backdoor attack mitigation, consistent with other scenarios. Among the remaining techniques, KD-FedUL shows limited unlearning capability with a success attack rate of around 15\%, attributed to its dependency on an additional unlabeled dataset for post-training, leading to potential distribution discrepancies and lower model quality. 

In summary, Fast-FedUL dramatically reduces execution time while efficiently eliminating the influence of target clients and preserving the knowledge of untargeted clients.

\vspace{-10pt}
\subsection{Ablation Study}
We compare Fast-FedUL with three variants: Fast-FedUL-1, which substitutes the proposed client sampling with random client sampling; Fast-FedUL-2, which utilizes only gradient ascent as existing methods without considering skew mitigation; and Fast-FedUL-3, which applies skew mitigation for half of the communication rounds.
\begin{figure}[bt]
\begin{minipage}{0.48\textwidth}
    \centering
    \captionof{table}{Comparison of Fast-FedUL's variants.}
    \label{tbl:ablation}
     \resizebox{1.0\linewidth}{!}{%
    \normalsize
    \begin{tabular}{l cc cc}
    \toprule
    \multicolumn{1}{c}{} & \multicolumn{2}{c}{\textbf{Edge-case}} & 
    \multicolumn{2}{c}{\textbf{Pixel}} \\ 
    \textbf{Setting} & main & backdoor 
    & main & backdoor \\ \midrule
    \textit{Fast-FedUL \quad} & \multicolumn{1}{l}{\textbf{0.9737}} & 0.0027 & \multicolumn{1}{l}{\textbf{0.9847}} & 0.0018 \\ 
    \textit{Fast-FedUL-1 \quad} & \multicolumn{1}{l}{0.9689} & 0.1074 & \multicolumn{1}{l}{0.9654} & 0.1542 \\     
    \textit{Fast-FedUL-2 \quad} & \multicolumn{1}{l}{0.8051} & \textbf{0.0000} & \multicolumn{1}{l}{0.8143} & \textbf{0.0000} \\ 
    \textit{Fast-FedUL-3 \quad} & \multicolumn{1}{l}{0.8527} & 0.0013 & \multicolumn{1}{l}{0.8732} & 0.0007 \\ 
    \bottomrule
    \end{tabular}
    }
\end{minipage}%
\hfill %
\begin{minipage}{0.48\textwidth}
    \centering
    \centering
    \subcaptionbox{Edge-case.}[0.48\textwidth]{
        \includegraphics[width=0.48\textwidth]{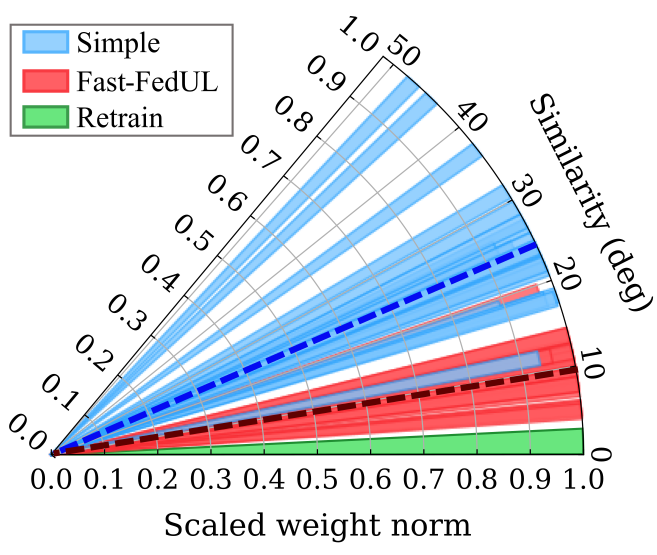}
    }\hfill %
    \subcaptionbox{Pixel.}[0.48\textwidth]{
        \includegraphics[width=0.48\textwidth]{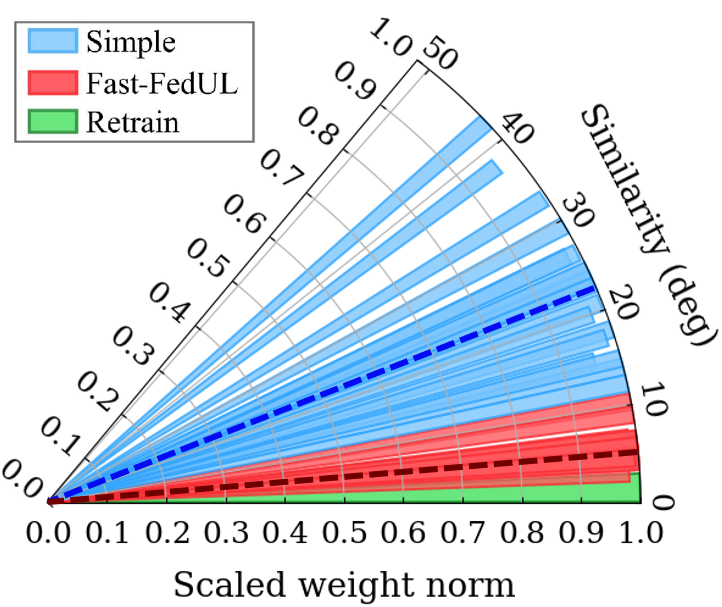}
    }
    \caption{Model deviations.}
    \label{fig:deviation}
\end{minipage}
\vspace{-12pt}
\end{figure}
\tabref{tbl:ablation} shows the performance comparison between the full model and its variants in two attack scenarios on MNIST dataset. Our model, Fast-FedUL, consistently outperforms other versions,  highlighting the advantages of our proposed techniques. In Fast-FedUL-1, the replacement of our proposed sampling with a random strategy slightly lowers main task performance but significantly reduces backdoor removal effectiveness. The removal of our proposed skew mitigation in Fast-FedUL-2 leads to a substantial 16-17\% quality drop, even though it effectively eliminates the backdoor threat. Meanwhile, the partial implementation of skew mitigation in Fast-FedUL-3 limits its ability to recover model utility, showing only a 4.76\% and 5.89\% quality improvement in the two scenarios, in contrast to the 16.86\% and 17.04\% improvement achieved by the full model. 
\vspace{-10pt}
\subsection{Qualitative Study}
\label{sec:exp_qualitative}

To highlight the role of our skew-resilient 
scheme, we compare the parameters of the model obtained with the skew-ignorant 
model (referred to as \textit{Simple}). The visualization of the parameter deviation histogram of the models is shown 
in \figref{fig:deviation}. The deviation is computed by $\theta=$ $\arccos 
\frac{w_u \dot{w}_r}{\left\|w_u\right\|\left\|w_r\right\|}$, where $w_u$ and $w_r$ are the last layer weight of the inspected model and the reference 
model, respectively. 
We observe that the model produced by Fast-FedUL is much closer to the ideal model than the skew-ignorant model. 
The mean angle deviation between the Fast-FedUL model and the retrained 
model is less than $10^{\circ}$, with a concentrated distribution and no 
deviation range being larger than $20^{\circ}$. The skew-ignorant model, on the 
other hand, has a mean deviation $2.6 \times$ higher than the Fast-FedUL model, 
and also shows a high deviation range. This confirms the need of considering 
the possible skew to the global model as well as the effectiveness of our 
skew-resilient strategy.

\section{Conclusion}
\label{sec:con}
In this paper, we proposed Fast-FedUL, a federated unlearning technique that reversely removes the historical gradients of a target client in an efficient and certified manner. Here, our novel angle is the streamlined sampling of the clients' updates to optimize the storage cost and an estimation of the model skew incurred by the deduction of the gradients. Based on a theoretical bound for this skew, we showed how to accumulate it in order to directly recover the utility of the global model. Since Fast-FedUL is training-free, it ultimately outperforms existing methods that rely on extra client-server communication or a separate knowledge distillation step. Experimental results obtained for backdoor attack scenarios on three benchmark datasets justified the advances of our techniques over SOTAs on model recovery, unlearning effectiveness and efficiency. In future work, we plan to explore further streaming strategies~\cite{nguyen2022detecting,nguyen2022model,zhao2021eires,nguyen2023example,nguyen2017retaining}.

%\bibliographystyle{splncs04}
%\bibliography{../ref,../ref_h}

\appendix
\section{Appendix}

\subsection{Notation Summary}

\begin{table}[H]
\centering
%\caption{Notation Summary}
%\label{tlb:symbols}
\scriptsize
\begin{tabular}{c p{5.5cm}}
\toprule
\textbf{Symbols} & \textbf{Definition} \\
\midrule
$S$ & central server $S$\\
$\{e_1,e_2,...,e_N\}$ & set of all clients\\
$\{D_1, D_2,..., D_N\}$ & set of all datasets of clients\\
$\mathcal{M}_t$ & global model at training round $t$ \\
$\Delta\mathcal{M}^i_t$ & update of client $e_i$ at round $t$\\
$\alpha$ & Lipschitz coefficient\\
$\epsilon_t^i$ & local skew for client $e_i$ at training round $t$\\
$\Delta_t = \mathcal{M}^*_t - \mathcal{M}_t$ & Difference between re-trained global model and original global model\\
$tr(X)$ & trace of matrix $X$\\
\bottomrule
\end{tabular}
%\vspace{-10pt}
\end{table}

\subsection{End-to-end Federated Learning and Unlearning Process}

Algorithm \ref{alg:ALG1} outlines our federated unlearning process integrated into an ongoing continuous federated learning (FL) pipeline. The FL process initiates by initializing the global model (Line 1) and then proceeds to iteratively distribute its training across client devices (Line 2-10). Within each training round, a subset of clients is efficiently selected using our sampling methods (Line 3). Subsequently, the chosen clients retrieve the global model (Line 5), locally train it with their respective data for a set number of iterations (Line 6-7), and transmit their local model gradients to the server (Line 8) for aggregation (Line 9). The updates from the sampled clients in each round are aggregated and stored at the server (Line 10), facilitating the federated unlearning process. Upon receiving an unlearning request from a user client $e_k$ (Line 11), our proposed efficient unlearning technique (as detailed in Alg. 1) removes their contributions up to the current round from the global model; and the client is excluded from subsequent training steps.

\begin{figure}[!h]
	\begin{minipage}[t]{1.0\linewidth}
	\removelatexerror
	\begin{algorithm}[H]
		\footnotesize
		\caption{FL with Unlearning}
		\label{alg:ALG1}    
		\SetKwProg{Fn}{function}{}{end}
		\SetKwFunction{computeUnlearn}{accumSkew}
		
		\SetKwInOut{Input}{input}
		\SetKwInOut{Output}{output}  
		\BlankLine
		\Input{	Clients $E = \{e_1, ..., 
			e_N\}$ with local data $\{D_1, ..., D_N\}$;\\  number of training iterations $T_t$; \\ number of sampled clients $m$                \\} 
		\Output{Global model $\mathcal{M}$}
			
		% \BlankLine
         Initialize the global model $\mathcal{M}_0$
            
		\For(\tcp*[f]{\scriptsize For each iteration}){$t \in \{1,\dots, 
		T\}$}{
			Sample $C^t$ from $E$\;
			\For(\tcp*[f]{\scriptsize For each sampled client}){ $e_i \in C^t$}{
                Download $\mathcal{M}_{t-1}$ from server to $\mathcal{M}_{t-1}^i$
                
                \For(\tcp*[f]{\scriptsize For each epoch}){
			$k\in \{0,\ldots,R-1\}$} {
			$\Delta\mathcal{M}^{i}_{t}= \mathit{local\_train}(\mathcal{M}^{i}_{t-1}, \mathcal{D}_i)$\;}    					
			Send $\Delta \mathcal{M}^{i}_{t}$ to server\;
			}
		
		$\mathcal{M}_{t}=\mathcal{M}_{t-1} + Agg(\{\Delta \mathcal{M}_{t-1}^i \mid e_i \in C^t\})$ 
  
        Store the updates from sampled clients $C^t$
  
        \If{unlearning request of user $e_k$}{$\mathcal{M}_t$ = \textit{unlearning}($\mathcal{M}_t, e_k, \alpha$) \tcp*[f]{\scriptsize Unlearn the target client from the model if requested}} 
            
		}
		
		\KwRet{$\mathcal{M}_{T}$}\; 
		\end{algorithm}
	\end{minipage}	

    \begin{minipage}[t]{1.0\linewidth}
	\removelatexerror
	\begin{algorithm}[H]
		\footnotesize
		\caption{Unlearning Process.}
		\label{alg:ALG2}    
		\SetKwProg{Fn}{function}{}{end}
		\SetKwFunction{computeUnlearn}{accumSkew}
		\SetKwInOut{Input}{input}
		\SetKwInOut{Output}{output}  
		\BlankLine
		\Input{Central server $S$; original global model $\mathcal{M}_{T}$; 
			clients $E = \{e_1, ..., 
			e_N\}$ with data $\{D_1, ..., D_N\}$; 
			target client 
			$e_u$; number of iterations $T$; number of sampled clients N, 
			Lipschitz coefficient 
			$\alpha$.} 
		\Output{Global model after unlearning $\mathcal{M}^*_{T}$.}

		$\Delta'_{0}=0$; {\scriptsize // Initialize the model difference}

		\For({\scriptsize // Compute the model difference iteratively}){$t\in \{1,\ldots 
		T\}$}    {
		$\Delta'_{t} = (1 + \alpha)\Delta'_{t-1} + 
		\frac{1}{N(N-1)}\displaystyle\sum_{e_i \in 
		C_F^{t-1}} \Delta \mathcal{M}^i_{t-1} - \textstyle\frac{1}{N} \Delta \mathcal{M}^u_{t-1}$\;
		}
		$\mathcal{M}'_T = \mathcal{M}_{T} + \Delta'_{T}$\;
		\KwRet{$\mathcal{M}'_{T}$}\; 
		\end{algorithm}
	\end{minipage}
\end{figure}

\subsection{Extended Experiments}

\sstitle{End-to-end Comparison} \tabref{tab:end-to-end-pixel} reports the accuracy of the global on the \textit{main task} and the \textit{backdoor task} using the unlearning techniques against \textit{pixel backdoor attack}. Similar to the \textit{edge-case backdoor} scenario, our approach restores the model's efficacy in the main task and entirely neutralizes the attack's impact. Following Fast-FedUL's unlearning process, the global model maintains high accuracy levels of 98.47\%, 94.13\%, and 87.52\% across the three datasets for the main task, which are nearly identical to the retraining model's performance. Furthermore, our technique efficiently mitigates the attack's threat, as evidenced by a success attack rate of less than 0.1\% across all three datasets.

\begin{table*}[!h]
    \centering
      \caption{End-to-end comparison with \textbf{pixel backdoor}.}
    \label{tab:end-to-end-pixel}
    % \vspace{-.5em}
    \resizebox{1.0\linewidth}{!}{%
    \scriptsize
      \begin{tabular}{l cc cc cc 
    cc cc cc cc }
    \toprule
     & \multicolumn{2}{c}{\textbf{Pre-unlearned}} & 
    \multicolumn{2}{c}{\textbf{Retrain}} & 
    \multicolumn{2}{c}{\textbf{Fast-FedUL}} & 
    \multicolumn{2}{c}{\textbf{FedEraser}} & 
    \multicolumn{2}{c}{\textbf{CDP-FedUL}} & 
    \multicolumn{2}{c}{\textbf{PGA-FedUL}} & 
    \multicolumn{2}{c}{\textbf{KD-FedUL}} \\ 
    \textbf{Dataset} & main & 
    backdoor & main & 
    backdoor & main & 
    backdoor & main & 
    backdoor & main & 
    backdoor & main & 
    backdoor & main & 
    backdoor \\ \midrule
    \textbf{MNIST} & \multicolumn{1}{l}{0.9878} & 0.9220 & 
    \multicolumn{1}{l}{0.9875} & 0.0036 & \multicolumn{1}{l}{0.9847} & 0.0018 
    & \multicolumn{1}{l}{0.9878} & 0.0036 & \multicolumn{1}{l}{0.9878} & 
    0.9056 & \multicolumn{1}{l}{0.8092} & 0.0346 & \multicolumn{1}{l}{0.9876} 
    & 0.0780 \\ 
    \textbf{CIFAR10} & \multicolumn{1}{l}{0.9657} & 0.8913 & 
    \multicolumn{1}{l}{0.9520} & 0.0092 & \multicolumn{1}{l}{0.9413} & 0.0132 
    & \multicolumn{1}{l}{0.9474} & 0.0096 & \multicolumn{1}{l}{0.9463} & 
    0.7952 & \multicolumn{1}{l}{0.7683} & 0.0273 & \multicolumn{1}{l}{0.9467} 
    & 0.0529 \\ 
    \textbf{OCTMNIST} & \multicolumn{1}{l}{0.8837} & 0.8869 & 
    \multicolumn{1}{l}{0.8847} & 0.0083 & \multicolumn{1}{l}{0.8752} & 0.0073 
    & \multicolumn{1}{l}{0.8551} & 0.0240 & \multicolumn{1}{l}{0.8810} & 
    0.8729 & \multicolumn{1}{l}{0.7844} & 0.0137 & \multicolumn{1}{l}{0.8550} 
    & 0.0475 \\ \bottomrule
    \end{tabular}%
   }   
\end{table*}

\begin{figure}[!h]
    \centering
    \begin{subfigure}{0.49\linewidth}
    \centering
    \includegraphics[width=0.8\linewidth]{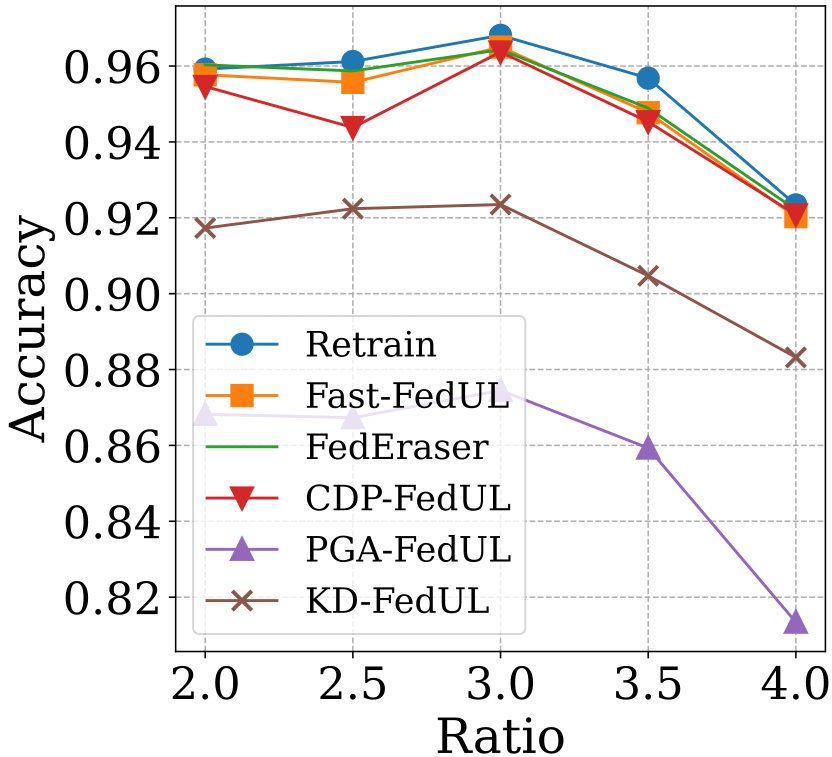}
    \caption{Main task $\mathcal{T}_m$}
    \end{subfigure}
    \hfill%
    \begin{subfigure}{0.49\linewidth}
    \centering
    \includegraphics[width=0.8\linewidth]{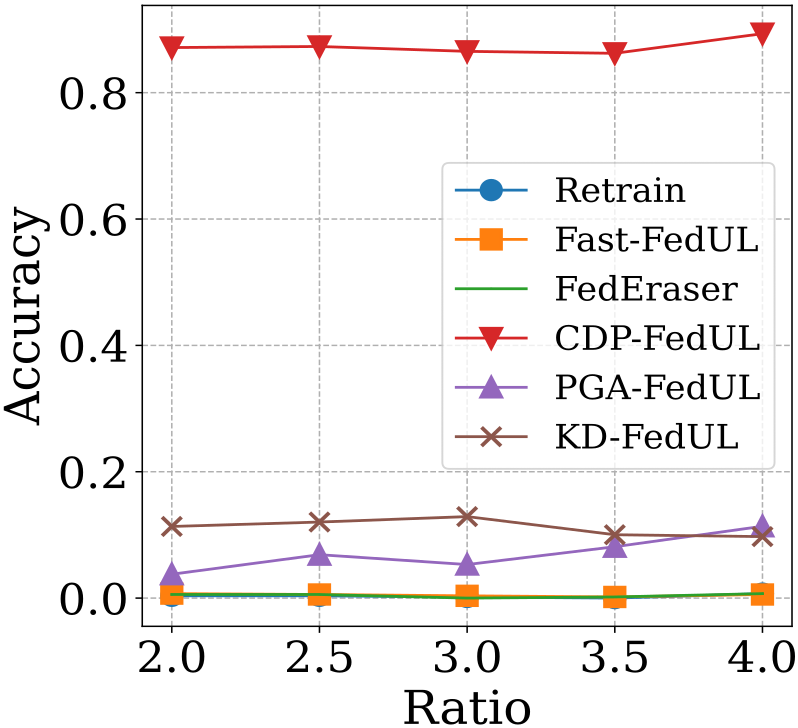}
    \caption{Backdoor task $\mathcal{T}_b$}
    \end{subfigure}

    \caption{Robustness against non-IID data (\textit{pixel backdoor}).}
    \label{fig:pixel_nonIID}
\end{figure}

\sstitle{Robustness to Data Distribution}
\figref{fig:pixel_nonIID} presents the resilience of the methods when confronted with non-IID data in the context of the \textit{pixel backdoor attack}. Similar to the situation in the \textit{edge-case attack} scenario, all approaches experience a decline in model effectiveness on the main task as the non-IID ratio increases. In this context, our Fast-FedUL and FedEraser techniques remain standout performers, demonstrating comparable quality to that of the retraining model. Conversely, PGA-FedUL exhibits significant vulnerability in non-IID scenarios, whereas CDP-FedUL consistently performs inadequately in mitigating backdoor attacks, as observed across various scenarios.

\sstitle{Hyper-parameter Sensitivity}
We explore the effect of the Lipschitz coefficient 
$\alpha$ on the performance of the model, using the MNIST dataset with $\alpha$ 
ranging from 0.01 to 0.11. For each $\alpha$, we report the accuracy of the 
main task and backdoor task of the model after each unlearning iteration. 

\begin{figure}[!h]
% \vspace{-1.5em}
    \centering
    \begin{subfigure}{0.49\linewidth}
    \centering
    \includegraphics[width=0.8\linewidth]{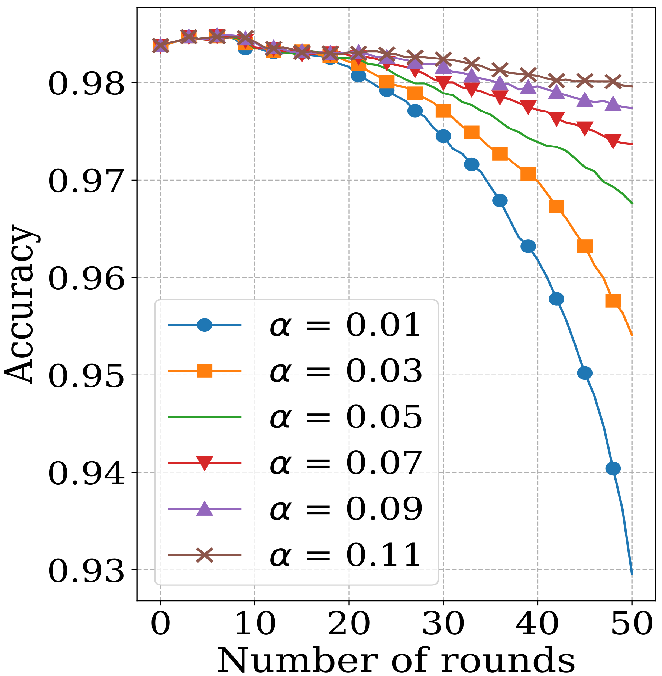}
    \caption{Main task $\mathcal{T}_m$}
    \end{subfigure}
    \hfill%
    \begin{subfigure}{0.49\linewidth}
    \centering
    \includegraphics[width=0.8\linewidth]{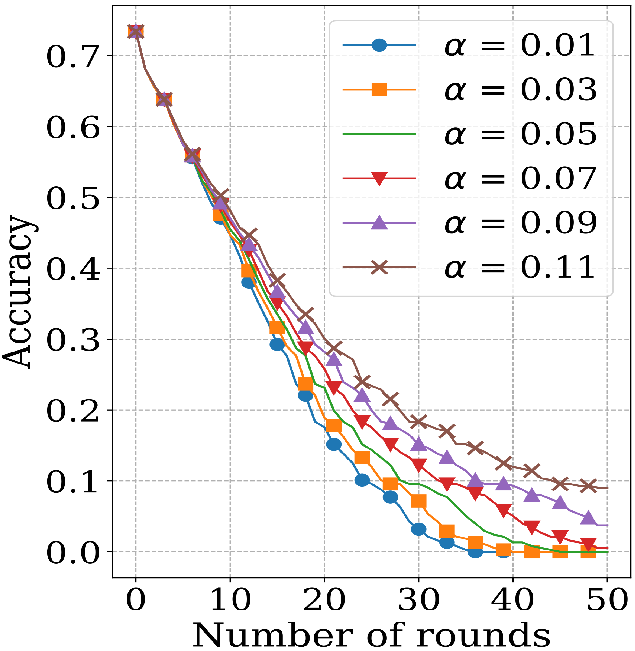}
    \caption{Backdoor task $\mathcal{T}_b$}
    \end{subfigure}
    % \vspace{-.5em}
    \caption{Effects of Lipschitz coef. $\alpha$ (\textbf{edge-case backdoor}).}
    \label{fig:hyperparameter_edge}
    % \vspace{-1.5em}
\end{figure}

\begin{figure}[!h]
    \centering
    \begin{subfigure}{0.49\linewidth}
    \centering
    \includegraphics[width=.8\linewidth]{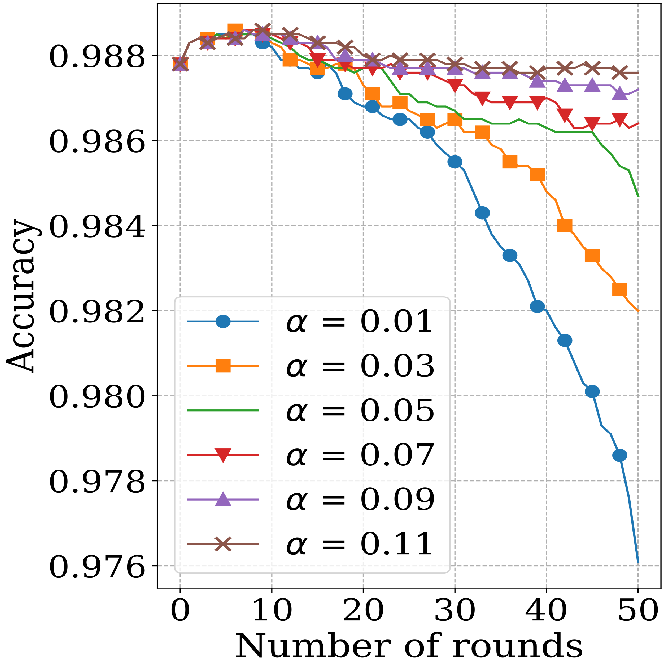}
    \caption{Main task $\mathcal{T}_m$}
    \end{subfigure}
    \hfill%
    \begin{subfigure}{0.49\linewidth}
    \centering
    \includegraphics[width=.8\linewidth]{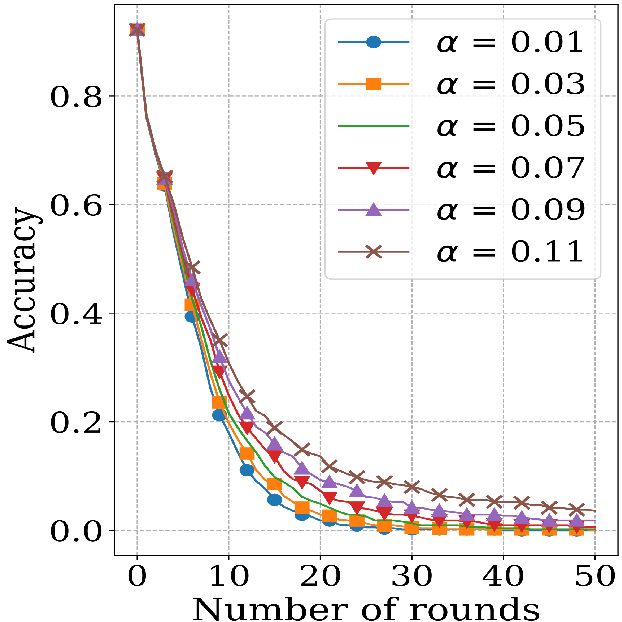}
    \caption{Backdoor task $\mathcal{T}_b$}
    \end{subfigure}

    \caption{Effects of Lipschitz coeff. $\alpha$ (\textbf{pixel backdoor}).}
    \label{fig:hyperparameter_pixel}
\end{figure}

The results in \figref{fig:hyperparameter_edge} indicate that a small change of $\alpha$ can lead to a 
considerable change in the final model's accuracy on both tasks. This is due to 
the skew in the model accumulating over the iterations of the unlearning 
process. Based thereon, we 
recommend the coefficient to be set in the range of 0.05 to 0.1.

\figref{fig:hyperparameter_pixel} depicts the sensitivity of our technique to the Lipschitz coefficient $\alpha$ under the \textit{pixel backdoor attack} scenario. This outcome aligns with observations from the \textit{edge-case backdoor attack}, indicating that the coefficient should ideally fall within the range of 0.05 to 0.1.

\subsection{Theoretical Remarks}

\begin{lemma} Let $\zeta_1, \zeta_2, \ldots, \zeta_N$ be vectors in $\mathbb{R}^d$ and $w_1, w_2, \ldots, w_N$ be non-negative numbers and $\sum_{i=1}^N w_i=1$, $C$ be a proper sampling. If $v \in \mathbb{R}^N$ is such that
\begin{equation}
\label{eqn:cond}
\mathbf{P}-p p^{\top} \preceq \operatorname{Diag}\left(p_1 v_1, p_2 v_2, \ldots, p_N v_N\right)
\end{equation}
then
$$
\mathbb{E}\left[\left\|\sum_{e_i \in C} \frac{w_i \zeta_i}{p_i}-\sum_{i=1}^N w_i \zeta_i\right\|^2\right] \leq \sum_{i=1}^N w_i^2 \frac{v_i}{p_i}\left\|\zeta_i\right\|^2,
$$
\noindent where $\mathbb{E}$ is the expectation taken over C. 
\end{lemma}

\noindent
Applying the lemma to Eq. 4 in main text, with $w_i = \frac{1}{N}$ and $\zeta_i = \Delta \mathcal{M}_t^i$, we have Eq. 5 in main text.

\subsubsection{Proof 1 - Optimal choice for $v_i$}
\begin{proof}
    From condition (\ref{eqn:cond}), we have:
    $$
    D = \operatorname{Diag}\left(p_1 v_1, p_2 v_2, \ldots, p_N v_N\right) - (\mathbf{P}-p p^{\top}) \succeq 0
    $$
    It is equivalent to $\forall z \in \mathbb{R}^N$:
    $$
    z^{\top}Dz \geq 0
    $$
    Consider $e_i = [0, 0,..., 1, 0,.., 0] \in \mathbb{R}^N$, where only $i$-th element of $e_i$ equals to $1$. Then we have: 
    $$
    p_i(v_i - 1 + p_i) = e_i^{\top}De_i \geq 0
    $$
    It implies that $v_i \geq 1 - p_i$
\end{proof}

\subsubsection{Proof 2 - Optimal bound for objective function}
\begin{proof}
    Our proof technique can be seen as an extended version of that in \cite{horvath2019nonconvex}. Let $1_{i \in C}=1$ if $i \in C$ and $1_{i \in C}=0$ otherwise. Likewise, let $1_{i, j \in C}=1$ if $i, j \in C$ and $1_{i, j \in C}=0$ otherwise. Note that $\mathrm{E}\left[1_{i \in C}\right]=p_i$ and $\mathrm{E}\left[1_{i, j \in C}\right]=p_{i j}$. Next, let us compute the mean of $X:=\sum_{i \in C} \frac{\Delta \mathcal{M}^i_t}{p_i}$ :
$$
\mathrm{E}[X]=\mathrm{E}\left[\sum_{i \in C} \frac{\Delta \mathcal{M}^i_t}{p_i}\right] = \sum_{i=1}^N \frac{\Delta \mathcal{M}^i_t}{p_i} \mathrm{E}\left[1_{i \in C}\right]=\sum_{i=1}^N \Delta \mathcal{M}^i_t %=\tilde{\zeta}
$$
Let $\boldsymbol{A} \in \mathbb{R}^{n \times n}$ be a matrix where $A_{ij}=tr\left(\frac{{\Delta \mathcal{M}^i_t}^{\top}}{p_i}\frac{\Delta \mathcal{M}^j_t}{p_j}\right)$, and let $e$ be the vector of all ones in $\mathbb{R}^N$. We now write the variance of $X$ in a form which will be convenient to establish a bound:
$$
\begin{aligned}
\mathrm{E}\left[\|X-\mathrm{E}[X]\|^2\right] & =\mathrm{E}\left[\|X\|^2\right]-\|\mathrm{E}[X]\|^2 \\
& =\mathrm{E}\left[\left\|\sum_{i \in C} \frac{\Delta \mathcal{M}^i_t}{p_i}\right\|^2\right]-\|\sum_{i=1}^N \Delta \mathcal{M}^i_t\|^2 \\
& =\mathrm{E}\left[\sum_{i, j} \boldsymbol{A}_{ij} 1_{i, j \in C}\right]-\|\sum_{i=1}^N \Delta \mathcal{M}^i_t\|^2 \\
& =\sum_{i, j} p_{i j} \boldsymbol{A}_{ij} -\sum_{i, j} tr\left( {\Delta \mathcal{M}^i_t}^{\top} \Delta \mathcal{M}^j_t \right)\\
& =\sum_{i, j}\left(p_{i j}-p_i p_j\right) \boldsymbol{A}_{ij} \\
& =e^{\top}\left(\left(\boldsymbol{P}-p p^{\top}\right) \circ \boldsymbol{A}\right) e .
\end{aligned}
$$
Since, by (\ref{eqn:cond}), we can further bound
\begin{align*}
e^{\top}\left(\left(\boldsymbol{P}-p p^{\top}\right) \circ  \boldsymbol{A}\right) e \leq e^{\top}\left(\boldsymbol{D i a g}(p \circ v) \circ \boldsymbol{A}\right) e =\sum_{i=1}^N p_i v_i \boldsymbol{A}_{ii}
\end{align*}
From those, we have:
\begin{equation}
    \label{eqn:proof2_equality}
    \mathrm{E}\left[\|X-\mathrm{E}[X]\|^2\right] \leq \sum_{i=1}^N p_i v_i \boldsymbol{A}_{ii} = \sum_{i=1}^N \frac{v_i}{p_i}\left\|\Delta \mathcal{M}^i_t\right\|^2
\end{equation}
Consider the case of independent sampling, then $\forall i \neq j: p_{ij} = p_ip_j$. It is equivalent to:
$$
\boldsymbol{P}-p p^{\top}=\boldsymbol{D i a g}(p \circ (1-p))
$$
For the optimal choice of $v_i$, the equality in (\ref{eqn:proof2_equality}) holds for independent sampling.
\end{proof}

\subsubsection{Proof 3 - Solution for optimal sampling} 
By Lemma 1, the independent sampling is optimal. In addition, for independent sampling, (\ref{eqn:proof2_equality}) holds as equality. We have:
\begin{align*}
\alpha_{C}:= \mathrm{E}\left[\sum_{i=1}^N \frac{1-p_i}{p_i}\left\|\Delta \mathcal{M}^i_t\right\|^2\right] =\mathrm{E}\left[\sum_{i=1}^N \frac{1}{p_i}\left\|\Delta \mathcal{M}^i_t\right\|^2\right] - \mathrm{E}\left[\sum_{i=1}^N\left\|\Delta \mathcal{M}^i_t\right\|^2\right]
\end{align*}
The optimal probabilities are obtained by minimizing $\alpha_{C}$ w.r.t. $\left\{p_i\right\}_{i=1}^N$ subject to the constraints $0 \leq p_i \leq 1$ and $m \geq b=\sum_{i=1}^N p_i$.

\begin{proof}
This proof uses an argument similar to that in the proof of Lemma 2 in \cite{horvath2019nonconvex} (Horvath \& Richtarik, 2019). The Lagrangian of our optimization problem is given by:
\begin{multline*}
L\left(\left\{p_i\right\}_{i=1}^N,\left\{\lambda_i\right\}_{i=1}^N,\left\{u_i\right\}_{i=1}^N, y\right) =\alpha_{C}\left(\left\{p_i\right\}_{i=1}^N\right)-\sum_{i=1}^N \lambda_i p_i
\\
- \sum_{i=1}^N u_i\left(1-p_i\right) -y\left(m-\sum_{i=1}^N p_i\right) .
\end{multline*}
Since all constraints are linear and the support of $\left\{p_i\right\}_{i=1}^N$ is convex, the KKT conditions hold. Therefore, the following solution is deduced from the KKT conditions:
$$
p_i= \begin{cases}\frac{(m+l-N)*\left\|\Delta \mathcal{M}^i_t\right\|}{\sum_{j=1}^l\left\|\Delta \mathcal{M}^{(j)}_t\right\|}, & \text { if } \left\|\Delta \mathcal{M}_t^i \right\| <\left\|\Delta \mathcal{M}_t^{(l+1)}\right\| \\ 1, & \text { otherwise } \end{cases}
$$
where $\left\|\Delta \mathcal{M}^{(j)}_t\right\|$ is the $j$-th largest value among the values $\left\|\Delta \mathcal{M}^1_t\right\|$,$\left\|\Delta \mathcal{M}^2_t\right\|$, $\ldots$,$\left\|\Delta \mathcal{M}^N_t\right\|; l$ is the largest integer for which $0<m+l-N \leq \frac{\sum_{i=1}^l\left\|\Delta \mathcal{M}^{(i)}_t\right\|}{\left\|\Delta \mathcal{M}^{(l)}_t\right\|}$.
\end{proof}

\subsubsection{Proof 4 - Optimal Sampling selects Attacked Clients.}

In the context of federated learning aimed at training a binary classifier, with malicious client $e_m$ and a random benign client $e_b$. We consider the case when $e_m$ and $e_b$ has benign data of same distribution (denote $\mathcal{D}_{clean}$ as dataset of this distribution), while $e_m$ has additionally a small set of backdoor data (denote $\mathcal{D}_{backdoor}$ with $|\mathcal{D}_{backdoor}|=\xi*|\mathcal{D}_{clean}|$).\\

Let $x, \tilde{x}$ be samples from $\mathcal{D}_{clean}$ (label '$0$') and $\mathcal{D}_{backdoor}$ (label '$1$'), respectively. At round $t$, global model $\mathcal{M}$ is sent to $e_b$ and $e_m$. Define $F$ as function that plays a role as Feature Extractor and $W$ is penultimate layer of $\mathcal{M}$, i.e. $\mathcal{M}(.) = softmax(W * F(.))$. The following condition can assure the selection on attacked clients:
\begin{lemma}
    \label{lem:3}
     For any function $F=[F^1;F^2;...;F^d]: \mathbb{R}^s \rightarrow \mathbb{R}_{\geq 0}^d$ such that each function $F^i$ is twice-differentiable and has continuous derivatives in an open ball $B$ with radius $\Delta x = \tilde{x}-x$ around the point $x$. 
    If Hessian Matrix of each function $F^i$ is semi-positive definite at any points between $x$ and $\tilde{x}$, and this condition satisfies $\forall i$:
    \begin{equation}
        \label{eqn:pseudo_convex}
        \Delta x^T \nabla F^i(x) > \frac{2}{\xi}*F^i(x) 
    \end{equation}
    then 
    \begin{equation}
        p_m > p_b
    \end{equation}
    where $p_m, p_b$ are probabilities for saving client $e_m$ and $e_b$, respectively.
\end{lemma}

\begin{proof}
We have that $\forall i: F^i$ is twice-differentiable and has continuous derivatives in an open ball of radius $\Delta x$. Implement Multivariate Taylor's expansion for $F^i$ around point $x$, note that $\tilde{x} = x + \Delta x$: 
\begin{equation}
    F^i(\tilde{x}) = F^i(x) + \Delta x^T \nabla F^i(x) + \frac{1}{2} (\Delta x)^T (\nabla^2 F^i(x_0))(\Delta x)
\end{equation}
where $x_0$ is a point that lies between $x$ and $\tilde{x}$ and $\nabla^2 f(x_0)$ is the Hessian of $f$ evaluated at a point $x_0$.\\
Because Hessian Matrix of $F^i$ is semi-positive, $\frac{1}{2} (\Delta x)^T (\nabla^2 F^i(x_0))(\Delta x) > 0$. Combine with the condition (\ref{eqn:pseudo_convex}), we have $\forall i$:
\begin{equation}
    \label{eqn:cp_ftExtractor}
    F^i(\tilde{x}) > \frac{\xi + 2}{\xi} F^i(x)
\end{equation}

Back to our analysis on gradient, we first compute gradient on $W$ regards to $x$ and $\tilde{x}$. 

Update in one cell of $W$:\\
For benign client $e_b$:
$$
\Delta w_{rc} = - \eta \mathrm{E}\left[\frac{\partial\mathcal{L}(W, x; y_r)}{\partial w_{rc}} \right]
$$
For malicious client $e_m$:
\begin{multline*}
    \Delta w_{rc} = - \eta
    (\frac{|\mathcal{D}_{clean}|}{|\mathcal{D}_{clean}|+|\mathcal{D}_{backdoor}|}\mathrm{E}\left[\frac{\partial\mathcal{L}(W, x; y_r)}{\partial w_{rc}} \right] \\
    + \frac{|\mathcal{D}_{backdoor}|}{|\mathcal{D}_{clean}|+|\mathcal{D}_{backdoor}|}\mathrm{E}\left[\frac{\partial\mathcal{L}(W, \tilde{x}; y_r)}{\partial w_{rc}} \right]) 
\end{multline*}
\begin{multline*}
    = - \eta \frac{1}{1 + \xi} ( \mathrm{E}\left[\frac{\partial\mathcal{L}(W, x; y_r)}{\partial w_{rc}} \right] 
    + \xi \mathrm{E}\left[\frac{\partial\mathcal{L}(W, \tilde{x}; y_r)}{\partial w_{rc}} \right])
\end{multline*}
Note that:
$$
\frac{\partial\mathcal{L}(W, x; y_r)}{\partial w_{rc}} = (softmax(W*F(x))_r - y_r)*F^c(x)
$$
and 
$$
\sum_r{L(x)_{rc}} = F^c(x) \sum_r (softmax(WF(x))_r - y_r) = 0
$$
where $L(x)_{rc} = \partial\mathcal{L}(W, x; y_r) / \partial w_{rc}$.\\

Then square of L2-norm for updates on $e_b$ and $e_m$ are respectively shown as:
$$
||\Delta_{b} W||^2 = 2\eta^2 \sum_c (\mathrm{E}\left[L(x)_{0c} \right])^2
$$
and 
$$
||\Delta_{m} W||^2 = \frac{2\eta^2}{(1+\xi)^2}\sum_c (\mathrm{E}\left[L(x)_{0c} \right] + \xi\mathrm{E}\left[L(\tilde{x})_{0c} \right])^2
$$
From that, we have:
\begin{multline}
    ||\Delta_{m} W||^2 - ||\Delta_{b} W||^2 = \frac{2\eta^2\xi}{(1+\xi)^2}\sum_c(\mathrm{E}\left[L(\tilde{x})_{0c} \right] \\
    - \mathrm{E}\left[L(x)_{0c} \right])(\xi\mathrm{E}\left[L(\tilde{x})_{0c} \right] + (\xi + 2)\mathrm{E}\left[L(x)_{0c} \right])
\end{multline}

Since $x$ has label '0' and $\tilde{x}$ has label '1', $\mathrm{E}\left[L(x)_{0c} \right]=(softmax(W*F(x))_0 - 1)*F^c(x) < 0$ and $\mathrm{E}\left[L(\tilde{x})_{0c} \right]=softmax(W*F(\tilde{x}))_0*F^c(x) > 0$. Moreover, we have $softmax(W*F(\tilde{x}))_0 > \mu > softmax(W*F(x))_1$, so:
\begin{multline}
    \xi\mathrm{E}\left[L(\tilde{x})_{0c} \right] + (\xi + 2)\mathrm{E}\left[L(x)_{0c} \right]
    > \mu (\xi \mathrm{E}\left[ F^c(\tilde{x}) \right] - (\xi+2)\mathrm{E}\left[ F^c(x) \right])
\end{multline}
Due to (\ref{eqn:cp_ftExtractor}), $\xi\mathrm{E}\left[L(\tilde{x})_{0c} \right] + (\xi + 2)\mathrm{E}\left[L(x)_{0c} \right] > 0$. Hence, we have $||\Delta_{m} W|| > ||\Delta_{b} W||$.\\

We consider these cases of $e_m$ and $e_b$:
\begin{enumerate}
    \item $m, b \in A^k$ or $m, b \notin A^k$, easily to see that $p_m > p_b$.
    \item $m \in A^k$ and $b \notin A^k$, $p_m=1 > p_b$.
    \item $m \notin A^k$ and $b \in A^k$, then $||\Delta_{b} W|| \geq ||\Delta W_{(l+1)}|| > ||\Delta W_{(l)}|| \geq ||\Delta_{m} W||$. (absurd)
\end{enumerate}
In all cases, we have $p_m > p_b$.
\end{proof}

\end{document}